\pdfoutput=1

\documentclass[11pt]{article}

\usepackage[final]{acl}

\usepackage{times}
\usepackage{latexsym}

\usepackage[T1]{fontenc}

\usepackage[utf8]{inputenc}

\usepackage{microtype}

\usepackage{inconsolata}

\usepackage{array,multirow,graphicx}
\usepackage{booktabs}
\usepackage{xcolor,colortbl,soul}
\usepackage{CJK}
\usepackage{adjustbox}
\usepackage{tabularx}
\usepackage[font=small,skip=-5pt]{subcaption}
\usepackage{enumitem}
\usepackage[ruled,vlined]{algorithm2e}
\usepackage[many]{tcolorbox}    	%
\usepackage{setspace}               %
\usepackage{multicol}               %
\usepackage{musicography}
\usepackage{tikz}
\usepackage{collcell}
\usepackage{xstring}

\newcommand*{\MinNumber}{0}%
\newcommand*{\MaxNumber}{60}%
\newcommand{\ApplyGradient}[1]{%
        \pgfmathsetmacro{\PercentColor}{100.0*(#1-\MinNumber)/(\MaxNumber-\MinNumber)}
        \hspace{-0.33em}\colorbox{red!\PercentColor!black}{}
}
\newcolumntype{R}{>{\collectcell\ApplyGradient}c<{\endcollectcell}}

\newcommand{\dataset}{SUnsET}

\newcommand{\sqa}{SQuALITY}
\newcommand{\las}{LexAbSumm}
\newcommand{\smh}{SummHay}
\newcommand{\ops}{ScholarQABench}

\newtcolorbox{sharp_box}{
    sharpish corners, %
    boxrule = 0pt,
    toprule = 4.5pt, %
    enhanced,
    fuzzy shadow = {0pt}{-2pt}{-0.5pt}{0.5pt}{black!35} %
}

\newtcolorbox{prompt_box}{
    enhanced,
    boxrule = 0pt,
    colback = sub,
    borderline west = {1pt}{0pt}{main}, 
    borderline west = {0.75pt}{2pt}{main}, 
    borderline east = {1pt}{0pt}{main}, 
    borderline east = {0.75pt}{2pt}{main}
}

\title{Unstructured Evidence Attribution for\\ Long Context Query Focused Summarization}

\author{Dustin Wright\thanks{Work partially completed during a research visit to University of Michigan.}\textsuperscript{\musSharp{}}\hspace{0.5cm}Zain Muhammad Mujahid\textsuperscript{\musSharp{}}\hspace{0.5cm}Lu Wang\textsuperscript{\musFlat{}} \\
\textbf{Isabelle Augenstein\textsuperscript{\musSharp{}}\hspace{0.5cm}David Jurgens\textsuperscript{\musFlat{}\musNatural{}}}\\
\textsuperscript{\musSharp{}}Department of Computer Science, University of Copenhagen\\ \textsuperscript{\musFlat{}}Department of Computer Science and Engineering, University of Michigan\\
\textsuperscript{\musNatural{}}School of Information, University of Michigan}

\begin{document}
\maketitle
\begin{abstract}

Large language models (LLMs) are capable of generating coherent summaries from very long contexts given a user query, and
extracting and citing evidence spans helps improve the trustworthiness of these summaries. 
Whereas previous work has focused on evidence citation with fixed levels of granularity (e.g. sentence, paragraph, document, etc.), we propose to extract \textit{unstructured} (i.e., spans of any length) evidence in order to acquire more relevant and consistent evidence than in the fixed granularity case. We show how existing systems struggle to copy and properly cite unstructured evidence, which also tends to be ``lost-in-the-middle''. 
To help models perform this task, we create the \textbf{S}ummaries with \textbf{Uns}tructured \textbf{E}vidence \textbf{T}ext dataset (\dataset{}), a synthetic dataset generated using a novel pipeline, which can be used as training supervision for unstructured evidence summarization. We demonstrate across 5 LLMs and 4 datasets spanning human written, synthetic, single, and multi-document settings that LLMs adapted with \dataset{} generate more relevant and factually consistent evidence with their summaries, extract evidence from more diverse locations in their context, and can generate more relevant and consistent summaries than baselines with no fine-tuning and fixed granularity evidence. We release \dataset{} and our generation code to the public.\footnote{\url{https://github.com/dwright37/unstructured-evidence-sunset}}
\end{abstract}

\section{Introduction}
\begin{figure}[t]
        \centering
        \includegraphics[width=0.98\linewidth]{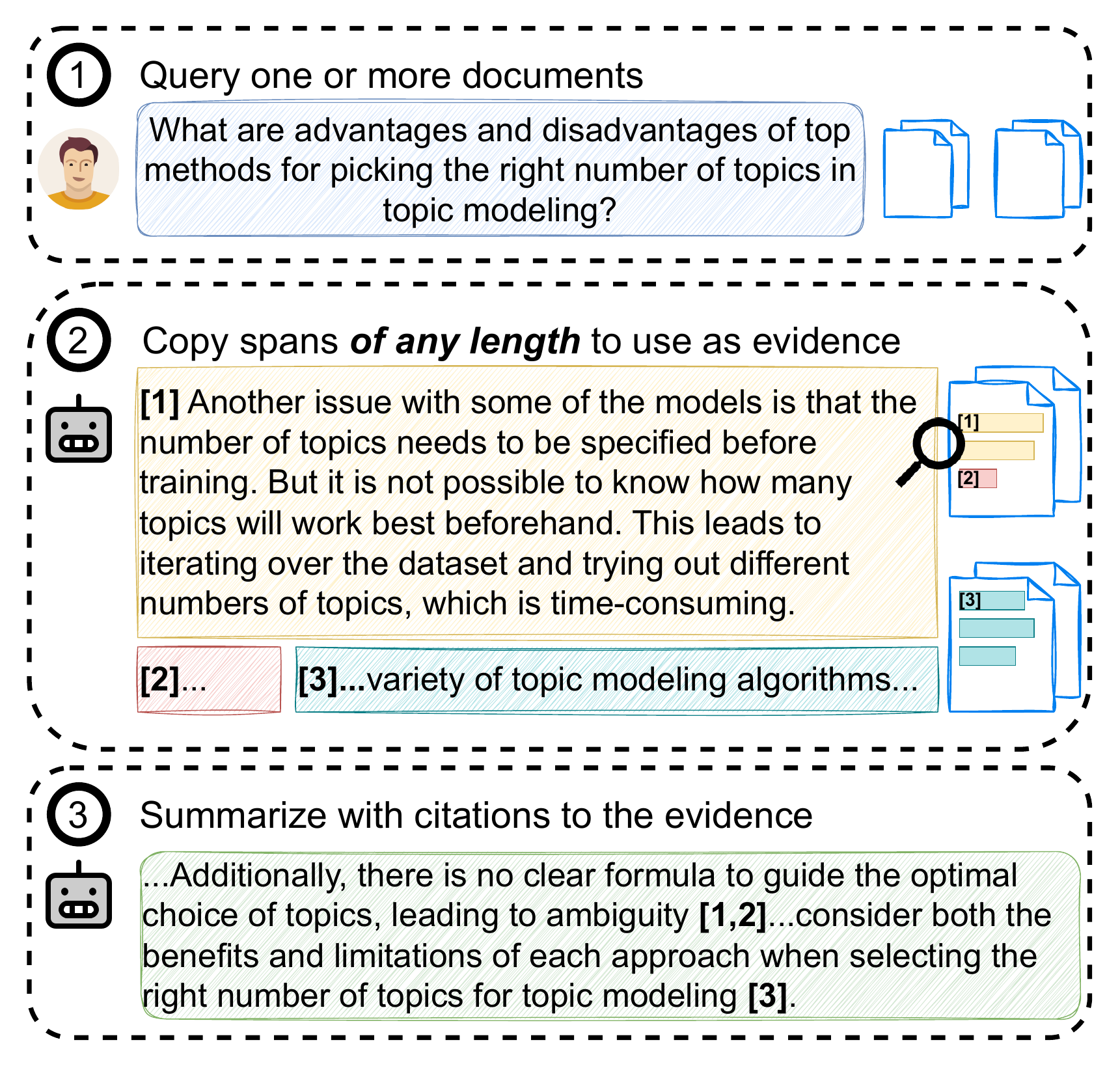}
        \caption{Summarization with \textit{unstructured} evidence requires a model to retrieve spans of any arbitrary length from the context to support individual sentences in the summary. Example given from Llama 3.1 8B trained on our dataset (\dataset{}).}
        \vspace{-15pt}
        \label{fig:fig1}
\end{figure}%

At the frontier of the capabilities of natural language processing (NLP) systems such as large language models (LLMs) is the ability to handle long contexts such as books and research papers, and summarize them based on queries~\cite{DBLP:journals/csur/KohJLP23,DBLP:journals/ijon/SuALPBL24,DBLP:journals/corr/abs-2004-05150,DBLP:journals/corr/abs-2403-05530}. %
While LLMs have progressed much on this~\cite{DBLP:journals/corr/abs-2404-16130}, people prefer traditional retrieval sources (e.g., search engines) for critical queries due to transparency and provenance~\cite{DBLP:journals/corr/abs-2411-17375}. Citing evidence in the summary addresses this, with prior work first segmenting the context into spans at fixed levels or granularity (e.g., sentences or documents, see \citealt{DBLP:journals/corr/abs-2311-03731}) and having models select evidence from among these segments to support the summary. As has been noted both in work on multi-document summarization~\cite{DBLP:conf/acl/ErnstSSABG0D24,DBLP:conf/acl/Xiao23} and automated fact checking~\cite{DBLP:conf/acl/WanCDLY20}, this approach is suboptimal for acquiring the most  \textit{salient} text in the context to support the summary, resulting in either too much or not enough information. In order to improve the precision of evidence in long-context query focused summarization (LCQFS), we propose to study \textit{unstructured} evidence citation, where any span of arbitrary length within the context can be used as evidence. 

In the unstructured evidence setup, a model must first copy spans from the context and subsequently use those spans as evidence in the summary (see \autoref{fig:fig1}). As we will show, simply prompting LLMs to perform this task with no other intervention leads to poor performance. Thus, we need to adapt models, e.g. through fine-tuning or in-context learning. For this, no suitable training data exist which consists of
examples of long documents, queries, summaries, and extracted evidence pointing to arbitrary spans in the documents. Based on the size and cost of other datasets for LCQFS~\cite{openscholar,DBLP:conf/emnlp/LabanFXW24,DBLP:conf/coling/SantoshAG24}, this would take an extensive amount of time, money, and expertise to create manually.

To address this, we present a synthetic dataset called the \textbf{S}ummaries with \textbf{Uns}tructured \textbf{E}vidence \textbf{T}ext dataset (\dataset{}). \dataset{} is generated using a novel pipeline, resulting in long documents paired with queries, summaries, and evidence spans.
We show that the data in \dataset{} are high quality and diverse, comparable to human written data. Using \dataset{}, we perform experiments 
across 5 models and 4 test datasets (including single- and multi-document, human and synthetic data), leading to the following findings: 1) for base LLMs with no fine tuning, extracting and citing unstructured evidence is challenging, and evidence is often lost-in-the-middle; 2) training on documents with shuffled structure (facilitated by \dataset{}) can help mitigate lost-in-the-middle, and 3) learning to cite unstructured evidence improves citation accuracy and coverage over fixed-granularity evidence, and additionally improves summary quality.

In sum, our contributions are:
\begin{itemize}[noitemsep]
    \item A synthetic dataset (\dataset{}) generated using a novel pipeline
    \item The first study on unstructured evidence citation for LCQFS, demonstrating that models adapted with \dataset{} produce higher quality evidence and summaries than baselines
    \item An analysis of and method to reduce the lost-in-the-middle problem with unstructued evidence
\end{itemize}

\section{Challenges in LCQFS}

LCQFS requires a model to be able to simultaneously ingest a large number of context tokens (possibly from multiple documents), retrieve and attend to relevant information in this context given a query, and integrate this information into a factually consistent and relevant summary. LLMs, with their increasingly large context sizes, have proven to be particularly adept at performing this task~\cite{DBLP:journals/tacl/ZhangLDLMH24,DBLP:journals/corr/abs-2404-16130,DBLP:journals/corr/abs-2408-14906}. Yet, a number of challenges remain, both in dealing with long contexts and with producing query-focused summaries~\cite{DBLP:conf/acl/LiWZZ24,DBLP:journals/corr/abs-2408-14906,DBLP:conf/acl/BaiLZL0HDLZHDTL24,DBLP:journals/tacl/LiuLHPBPL24,DBLP:conf/emnlp/0002IEBL23,DBLP:conf/acl/RavautSCJ24,DBLP:conf/emnlp/LabanFXW24,DBLP:journals/corr/abs-2411-17375, DBLP:journals/csur/JiLFYSXIBMF23,DBLP:conf/acl/ErnstSSABG0D24}. The main foci of our work are evidence attribution~\cite{DBLP:conf/emnlp/LabanFXW24,DBLP:journals/corr/abs-2411-17375,DBLP:journals/corr/abs-2311-03731,DBLP:conf/acl/ErnstSSABG0D24, DBLP:conf/acl/FierroAHCMNL24} and evidence being lost-in-the-middle~\cite{DBLP:journals/tacl/LiuLHPBPL24,DBLP:conf/acl/RavautSCJ24}, described next.

\begin{figure}[tb]
    \centering
    \begin{sharp_box}
    \fontsize{10pt}{10pt}\selectfont
            \textbf{Fixed-Granular Single Sentence Citation:} 
            
            \textsc{Summary Snippet:} ...[48] explains that the legend of the Ghost Ship is often told by space men as a cautionary tale....\\

            \textsc{Evidence:} [48] He had heard it spoken of in whispers by drunken space men and professional tellers of fairy tales.

            \noindent\rule{6.8cm}{0.4pt}

            \textbf{Unstructured Citation:} 
            
            \textsc{Summary Snippet:} ...he, like the ship's former crew, is doomed to wander in space, never able to return to Earth, a haunting reminder of what he has lost and what he can never have [2]...\\

            \textsc{Evidence:} [2] Doomed for all eternity to wander in the empty star-lanes, the Ghost Ship haunts the Solar System that gave it birth. And this is its tragedy, for it is the home of spacemen who can never go home again.
    \end{sharp_box}
    \caption{Examples of fixed-granular and unstructured evidence generated by models in our study. Fixed granular citations may include irrelevant or not enough information to support their citing sentences. Unstructured evidence allows for more flexible and precise evidence.}
    \vspace{-7pt}
    \label{fig:flexibility_example}
\end{figure}

\subsection{Evidence Attribution}
Improving the ability of LLMs to both generate relevant summaries and provide accurate attributions has the potential to help improve their usefulness, transparency, and trustworthiness. Recent work has started to explore this direction for LCQFS, including SummHay~\cite{DBLP:conf/emnlp/LabanFXW24} and OpenScholar~\cite{openscholar}. However, most works focus on fixed-granularity evidence (e.g., spans, sentences, paragraphs, or documents,~\citet{DBLP:journals/corr/abs-2311-03731}). Being able to flexibly cite evidence of any arbitrary length can lead to higher quality summaries which use \textit{precise} pieces of evidence from the context~\cite{DBLP:conf/acl/WanCDLY20,DBLP:conf/acl/ErnstSSABG0D24,DBLP:conf/acl/Xiao23}, as opposed to full documents which contain irrelevant information or individual sentences which may contain not enough information (see e.g., \autoref{fig:flexibility_example}). To the best of our knowledge, we provide a first study on unstructured evidence citation in LCQFS with LLMs.

\subsection{Lost-in-the-Middle}
LLMs suffer from positional preferences in their learned attention~\cite{DBLP:journals/tacl/LiuLHPBPL24}, oftentimes preferring early or late tokens in their context~\cite{DBLP:journals/corr/abs-2404-01430}. While this problem was originally demonstrated on retrieval-augmented-generation (RAG) tasks with explicit answers such as question answering, follow-up work has shown its persistence in more abstractive tasks such as summarization~\cite{DBLP:conf/acl/RavautSCJ24} and query focused multi-document summarization~\cite{DBLP:conf/emnlp/LabanFXW24}. A number of solutions have been proposed, most of which rely on manipulating either the positions of tokens in the context or the positional embeddings of LLMs in order to remove their intrinsic bias~\cite{DBLP:journals/corr/abs-2407-01100,DBLP:conf/acl/HePDSLQLWZZ24,DBLP:journals/corr/abs-2404-01430}. We explore and document this problem at the level of unstructured evidence citation, demonstrating how evidence is extracted unevenly across documents, and how this problem can be mitigated using purely synthetic data.

\section{Learning to Use Unstructured Evidence}
\label{sec:methods}

Our task is: given a query about a long input consisting of one or more documents, generate a response to the query which cites arbitrary length text spans from the input. 
This introduces challenges over the fixed-granularity case~\cite{DBLP:conf/emnlp/LabanFXW24,openscholar,DBLP:journals/corr/abs-2311-03731}, as targeted, precise evidence spans must be accurately copied from the context which are relevant and consistent with the summary sentences. While challenging, this can lead to summaries with more accurate and supportive evidence (\citealt{DBLP:conf/acl/ErnstSSABG0D24}).

\begin{figure}[tb]
    \centering
    \begin{sharp_box}
    \fontsize{10pt}{10pt}\selectfont
            \textbf{P1. Titles:}
            Generate $N$ unique titles of fiction and non-fiction documents.
            
            \textbf{P2. Document outline:} Given a title, generate an outline broken down into discrete sections.

            \textbf{P3. Queries, summaries, and evidence:} Given a document title and outline, generate 5 questions, 5 responses, and supporting passages that will be included in the document. 

            \textbf{P4. Document sections:} Generate each section of the document one at a time. Ensure that evidence passages are included verbatim. %

            \textbf{P5. Refinement:} For each $\langle$question, summary, evidence$\rangle$ tuple, refine the summary and evidence based on the document.
            
            \textbf{P6. Validation:} For each $\langle$question, summary, evidence$\rangle$ tuple, validate that the summary fully addresses the question, is faithful to the document, and includes inline attribution to evidence passages.
            
    \end{sharp_box}
    \caption{Six stage inductive data generation pipeline. The full prompts for each stage are given in Appendix \ref{sec:prompts} \autoref{fig:synth_title_prompt} - \autoref{fig:validation_prompt}.}
    \vspace{-7pt}
    \label{fig:dataset_generation}
\end{figure}

Large scale synthetic datasets are useful for fine-tuning task specific models at a lower cost than manual annotation~\cite{DBLP:journals/corr/abs-2409-02098,DBLP:conf/acl/HonovichSLS23, DBLP:conf/acl/WangKMLSKH23, DBLP:conf/iclr/ChenLYWGYTS0HJ24, DBLP:conf/iclr/XuSZG0FTLJ24}. To train LLMs to use unstructured evidence, we create \dataset{}, a synthetic dataset based on a novel inductive generation pipeline.
Training is performed using adapters~\cite{DBLP:conf/icml/HoulsbyGJMLGAG19} to improve unstructured evidence citation and mitigate the lost in the middle problem.  For the latter, previous work has shown that fine-tuning with data augmentation \citep[e.g., shuffling documents;][]{DBLP:journals/corr/abs-2404-01430} can help achieve this. Given this, we construct \dataset{} so that documents are modular: documents are broken down into discrete sections, so that data augmentation through shuffling document sections (thus shuffling global structure) is possible. We first present the inductive pipeline approach used to generate \dataset{}, followed by our two fine-tuning schemes.

\subsection{Generating \dataset{}}
 
Our pipeline generates long documents paired with queries, and summaries which address those queries. Each summary additionally includes citations which reference relevant text spans in the original document. We make several design decisions intended to overcome known problems in synthetic data generation, including the potential for low diversity~\cite{DBLP:conf/acl/HonovichSLS23,DBLP:conf/acl/WangKMLSKH23} and labeling errors~\cite{DBLP:conf/iclr/ChenLYWGYTS0HJ24}. This includes taking a six stage pipeline approach which generates synthetic data inductively, and validation steps which refine summaries, refine evidence, and reject bad summaries and evidence.

\begin{figure}[t]
    \centering
    \begin{sharp_box}
    \fontsize{10pt}{12pt}\selectfont

            \textbf{Example Document Snippet}\\
            Title: ``Writing the Unwritable''\\
            ...They demonstrate that while writing the unwritable is fraught with difficulty, it can also yield transformative insights that resonate profoundly with readers. \hl{Writing the unwritable requires a recognition of the limitations of language, and a willingness to push against those boundaries.} This requires not merely acceptance of silence or ambiguity but a bold declaration that some truths demand to be told, no matter how fraught the endeavor may be....

            \textbf{Example Query}\\
            What does it mean to write the unwritable, and what historical examples illustrate this concept?

            \textbf{Example Summary Snippet}\\
            To write the unwritable involves confronting and articulating subjects and experiences that resist verbal expression, often due to limitations of language, social taboos, and the impact of censorship [1][2][3].

            \textbf{Example Evidence Snippet}\\
            $[$1$]$ \hl{Writing the unwritable requires a recognition of the limitations of language, and a willingness to push against those boundaries.}
            
    \end{sharp_box}
    \caption{Snippets from a \dataset{} document.}
    \label{fig:data_sample}
\end{figure}

The full generation process is described in \autoref{fig:dataset_generation}, with prompts provided in Appendix \ref{sec:prompts}. Diversity in document topic and type is accomplished by first generating document titles which seed the subsequent steps. We inductively build up each document, starting with the queries, summaries, and evidence passages. When generating evidence, each evidence passage is assigned to a section in the document so that evidence can be distributed precisely.
The summaries, queries, and assigned evidence are then used as context to generate each section of the document one at a time. This makes documents modular, which we take advantage of during training to study lost-in-the-middle. Following this, the queries, summaries, and evidence are refined by using the final document as context. Finally, we filter out poor summaries and evidence by prompting to predict if the summaries fully address the query and are fully supported by the document (see \autoref{fig:data_sample} for an example). In total we generate 2,352 synthetic documents, giving us 11,309 $\langle$document, question, summary$\rangle$ tuples. 

\paragraph{Cost Comparison} Manually annotating data of the kind in \dataset{} is highly expensive, requiring annotators to read long sets of documents with long summaries and verifying the quality of the references. As a comparison, SQuALITY~\cite{DBLP:conf/emnlp/WangPCPB22} is a similar dataset to ours in terms of document and response size, and they paid Upwork workers \$13 to write each response, followed by \$8 to review each response in their data. As we generated 11,309 responses in SUnsET, this alone would have cost \$237,468. In contrast, generating SUnsET, including documents, questions, responses, and evidence, cost around \$200.

\begin{table}[t]
    \setlength{\tabcolsep}{2pt}
    \def\arraystretch{1.1}
    \centering
    \fontsize{7.8}{7.8}\selectfont
    \rowcolors{2}{gray!15}{white}
    \begin{tabular}{r c c c | c c c | c c c}
    \toprule 
    & \multicolumn{3}{ c |}{ \dataset{} } & \multicolumn{3}{c|}{ Non-Pipelined } & \multicolumn{3}{c}{ Title + Doc } \\
    \midrule
    
    \cellcolor{black!75}\color{white}Metric & Q & S & D  & Q & S & D & Q & S & D \\
    \cellcolor{black!75}\color{white}TTR & \textbf{0.75} & \textbf{0.84} & \textbf{0.82} &  0.67 & 0.80 & 0.35 & 0.63 & 0.78 & 0.35 \\
    \cellcolor{black!75}\color{white}Cos & \textbf{0.81} & \textbf{0.73} & \textbf{0.68} &  0.73 & 0.72 & 0.04 & 0.66 & 0.61 & 0.04 \\
    \cellcolor{black!75}\color{white}Len & \textbf{13.45} & \textbf{226.5} & \textbf{3767.4} & 9.85 & 23.79 & 474.8 & 10.21 & 24.45 & 433.8 \\


    \bottomrule 

    \end{tabular}
    \caption{Statistics and diversity metrics of synthetic data. Metrics are average type-token ratio (TTR) \citet{DBLP:journals/corr/abs-2307-04626}, embedding cosine distance (Cos), and average word length (Len). Columns differentiate between (Q)uestion, (S)ummary and (D)ocument metrics in each dataset. \textbf{Bold} is highest diversity across datasets.} 
    \label{tab:synthetic_stats}
\end{table}

\paragraph{Evaluation} We evaluate both the quality and diversity of data generated using this pipeline. %
For quality, we asked two independent annotators (NLP researchers unaffiliated with the project) three questions for 100 $\langle$question, summary, evidence$\rangle$ tuples: Q1) Does the summary address the question?; Q2) Is the summary well structured and organized; and Q3) Does the evidence fully support the summary? Annotators responded to each question with one of the following values: 1 - Not at all; 2 - Somewhat; 3 - Completely. We find that the data is very high quality, acquiring scores of 2.99 for Q1, 2.97 for Q2, and 2.90 for Q3, with an exact agreement rate of 93.67\% across all 300 annotations.

\begin{table}[t]
    \setlength{\tabcolsep}{2pt}
    \def\arraystretch{1.1}
    \centering
    \fontsize{7.8}{7.8}\selectfont
    \begin{tabular}{r c}
    \toprule 
    Dataset & Topic Diversity \\
    \midrule
    
    Non-Pipelined & \cellcolor{blue!20}0.506 \\
    Title + Doc & 0.356 \\
    \sqa{} (human, stories)& \cellcolor{blue!45}0.705 \\
   \las{} (human, legal text) & \cellcolor{blue!40}0.673 \\
    \ops{} (human, scientific docs) & \cellcolor{blue!44}0.695 \\
    \textbf{\dataset{}} & \cellcolor{blue!41}0.679 \\


    \bottomrule 

    \end{tabular}
    \caption{Topic diversity scores using the approach from \citet{DBLP:conf/eacl/TerragniFGTC21}. Shading indicates magnitude of diversity score.} 
    \label{tab:topic_diversity}
\end{table}

To validate \dataset{} diversity, we generate two baseline datasets. The first is generated by combining all the steps in \autoref{fig:dataset_generation} into one prompt, forcing the model to simultaneously perform all tasks to generate each example (called Non-Pipelined). The second includes a title generation step to seed each document (called Title + Doc, see \autoref{fig:baseline_prompt} in Appendix \ref{sec:prompts} for prompts). We compare each dataset using samples of 100 documents along lexical and semantic diversity metrics in \autoref{tab:synthetic_stats}. Further, in \autoref{tab:topic_diversity} we compare the topic diversity (following \citealt{DBLP:conf/eacl/TerragniFGTC21}) between these datasets, as well as three human-written datasets:
\textbf{\sqa} \cite{DBLP:conf/emnlp/WangPCPB22}, \textbf{\las{}} \cite{DBLP:conf/coling/SantoshAG24}, and \textbf{\ops} \cite{openscholar},  (see Appendix \ref{sec:topic_diversity}. Our approach generates longer documents with longer summaries than baseline non-pipelined approaches, which also tend to be much more diverse. Additionally, our pipeline produces documents with topic diversity similar to that of human written datasets.

\subsection{Training Complementary Adapters}

Previous work has demonstrated that altering the position embeddings of LLMs either directly or through fine-tuning can help to overcome positional biases~\cite{DBLP:conf/acl/HsiehCL0LKGRLKP24,DBLP:journals/corr/abs-2404-01430}. 
We design \dataset{} documents so that they are modular, having global coherence at the level of the full document and local coherence at the level of discrete sections. Given this, we experiment with position-aware and position-agnostic training in order to observe their impact on evidence selection and quality, as well as summary quality. For position-aware training, we concatenate all document sections together in their natural order to construct the context, while for position-agnostic training, we shuffle the document sections before concatenating them, thus randomizing the global structure of the position embeddings while maintaining the local structure. This gives us two adapters for each model in our experiments. The prompt we use for training is provided in Appendix \ref{sec:prompts} \autoref{fig:generation_prompt}, and all training is performed using supervised fine-tuning on \dataset{} data using LoRA~\cite{DBLP:conf/iclr/HuSWALWWC22}. In all cases we fine tune using the Huggingface Transformers implementation of LoRA~\cite{DBLP:conf/iclr/HuSWALWWC22} with a rank and $\alpha$ of 16 applied to all linear operators of each model.

\subsection{Summarizing with Unstructured Evidence}

To generate summaries with unstructured evidence, we use the prompt from~\citet{openscholar}, altering it to include unstructured evidence extraction as a first step. The full prompt is given in \autoref{fig:generation_prompt} in Appendix \ref{sec:prompts}. We use this prompt for both inference and supervised fine-tuning on \dataset{}. 
To deal with long contexts, we divide-and-conquer by chunking each document by the model's maximum token length, summarize each chunk, and finally summarize the summaries. Thus, the output for each $\langle$document, query$\rangle$ pair is a $\langle$summary, evidence\_list$\rangle$ pair containing the summary and a list of evidence text from the context. 

\section{Experiments and Results}

Our experiments focus on three research questions:
\begin{itemize}[noitemsep]
    \item \textbf{RQ1:} How well can LLMs extract and use unstructured evidence?
    \item \textbf{RQ2:} Is evidence lost-in-the-middle?
    \item \textbf{RQ3:} Does learning to cite unstructured evidence improve summary quality?
\end{itemize}

\begin{figure*}[t]
        \centering
        \includegraphics[width=0.98\linewidth]{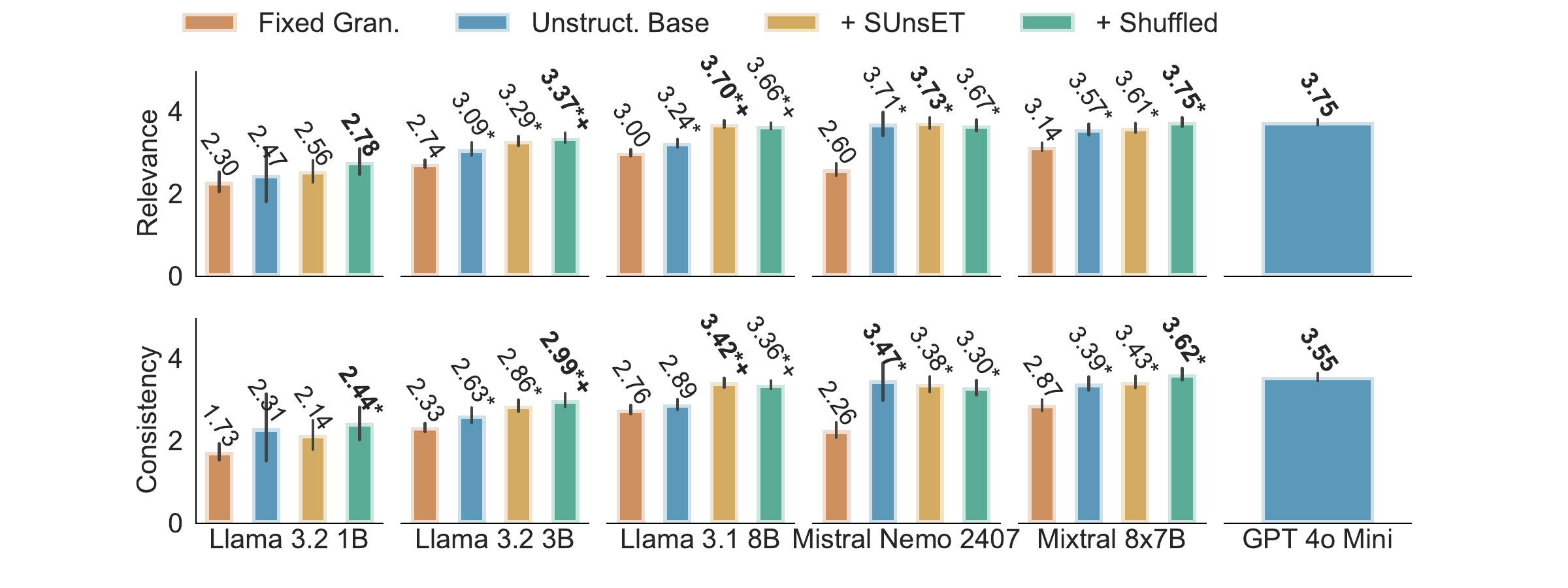}
        \caption{Average relevance and consistency of evidence texts with respect to their citation sentences measured using an autorater~\citep[DeepSeek-V3;][]{DBLP:conf/emnlp/LiuIXWXZ23} based on prompts which have previously undergone human evaluation for quality~\cite{DBLP:journals/corr/abs-2410-23463}. 
    \textbf{Bold} indicates best performance for a given model; ``*'' and ``+'' indicate statistical significance above the fixed granularity and non-fine-tuned unstructured baselines, respectively, based on non-overlapping 95\% confidence intervals.}
        \label{fig:citation_avg}
\end{figure*}

\paragraph{Test Data} We use four test datasets (full descriptions in Appendix \ref{sec:datasets_appendix}). These include three human written datasets, forcing models trained on \dataset{} to generalize beyond synthetic data. These are: \textbf{\sqa} (\citealt{DBLP:conf/emnlp/WangPCPB22}, short sci-fi novels, single document, average context length: 5,200 tokens); \textbf{\las{}} (\citealt{DBLP:conf/coling/SantoshAG24}, long legal documents, single document, average context length: 14,357 tokens); \textbf{\smh} (\citealt{DBLP:conf/emnlp/LabanFXW24}, synthetic conversations and news, multi-document, average haystack context length: 93,000 tokens); and \textbf{\ops} (\citealt{openscholar}, Computer Science research papers, multi-document, average context length: 16,341 tokens). We present here the average results from sampling evenly across datasets, results on individual datasets are presented in Appendix \ref{sec:individual_results}.

\paragraph{Models} We test Llama 3.2 1B, Llama 3.2 3B, Llama 3.1 8B~\cite{DBLP:journals/corr/abs-2407-21783}, Mistral Nemo 2407, and Mixtral 8x7B.\footnote{Huggingface model IDs are listed in Appendix \ref{sec:models_appendix} \autoref{tab:models_description}} We compare four settings for each LLM: base models with fixed granularity evidence (Fixed Gran.), base models with unstructured evidence citation (Unstruct. Base), training adapters on \dataset{} (+ SunSET), and training adapters on shuffled \dataset{} documents (+ Shuffled).  Additionally, we provide an upper bound estimate on performance using GPT 4o mini with no fine-tuning. 

\begin{table}[t]
    \setlength{\tabcolsep}{3.5pt}
    \def\arraystretch{1.0}
    \centering
    \fontsize{8}{8}\selectfont
    \begin{tabular}{l c c c}
    \toprule 
    Model & Exact Match & 50\% Match & \# Evidence\\
\midrule
Llama 3.2 1B &  ~0.0 & 35.71 & ~~~~14 \\
+ \dataset{} & ~~\textbf{7.69} & \textbf{43.26} & ~~208\\
+ Shuffle & ~~5.15 & 22.68 & ~~~~97\\\midrule
Llama 3.2 3B &  25.57 & \textbf{90.11} & 1345\\
+ \dataset{} & \textbf{52.77} & 85.62 & 3720\\
+ Shuffle & 32.99 & 74.07 & 2337 \\\midrule
Llama 3.1 8B &  43.93 & 83.12 & 3412\\
+ \dataset{} & \textbf{78.36} & \textbf{97.21} & 4690\\
+ Shuffle & 54.53 & 88.51 & 4684 \\\midrule
Mistral Nemo 2407 &  ~~5.48 & 66.13 & ~~310\\
+ \dataset{} & \textbf{82.20} & \textbf{97.29} & 2107\\
+ Shuffle & 72.38 & 95.76 & 1959 \\\midrule
Mixtral 8x7B &  ~~5.79 & 91.25 & 3452\\
+ \dataset{} & \textbf{33.82} & 90.47 & 4208\\
+ Shuffle & 29.29 & \textbf{90.74} & 4288\\
\midrule
\midrule
GPT-4o-mini & \textit{11.06} & \textit{96.32} & \textit{8159}\\
    \bottomrule %

    \bottomrule 

    \end{tabular}
    \caption{
    Evidence copy rates. We measure exact string match (i.e. when the evidence sentence \textit{exactly} appears in the context) as well as 50\% overlap between the extracted evidence and the longest common substring.} 
    \vspace{-7pt}
    \label{tab:hallucination_results}
\end{table}

\paragraph{Evaluation} 
We evaluate our models using autoraters~\cite{DBLP:journals/corr/abs-2411-15594,DBLP:conf/nips/ZhengC00WZL0LXZ23,DBLP:conf/emnlp/LiuIXWXZ23} along two dimensions. These dimensions are \textit{Relevance} and \textit{Consistency}. Given a source text, a target text, and optionally a query, \textit{Relevance} measures how well the target covers the main points of the source, as well as how much irrelevant or redundant information it contains. \textit{Consistency} measures to what degree the target contains any factual errors with respect to the source. Both scores are measured on a scale from 1-5 using DeepSeek-V3~\cite{liu2024deepseek}.\footnote{We validate the robustness of the ratings from DeepSeek-V3 in Appendix \ref{sec:evaluation_robustness}.} We use prompts which have been previously validated to correlate well with human annotations of relevance and consistancy (listed in Appendix \ref{sec:prompts} \autoref{fig:relevance_prompt} and \autoref{fig:consistency_prompt})~\cite{DBLP:journals/corr/abs-2410-23463}.

\begin{figure*}[t]
        \centering
        \includegraphics[width=0.98\linewidth]{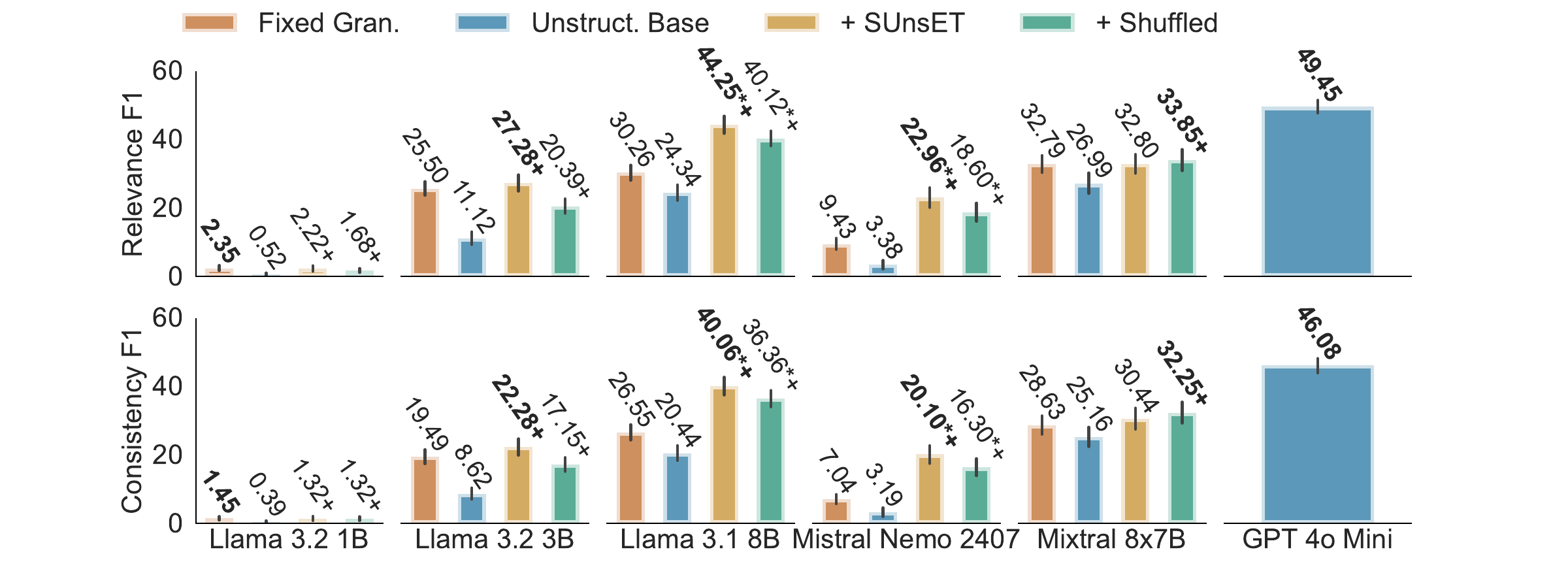}
        \caption{Relevance and consistency F1 scores.
    \textbf{Bold} best performance for a given model; ``*'' and ``+'' indicate statistical significance above the fixed granularity and non-fine-tuned unstructured baselines, respectively, based on non-overlapping 95\% confidence intervals.}
        \label{fig:citation_results}
\end{figure*}
\begin{figure*}[t]
    \centering
    \begin{subfigure}[t]{0.27\textwidth}
        \centering\includegraphics[width=0.88\linewidth]{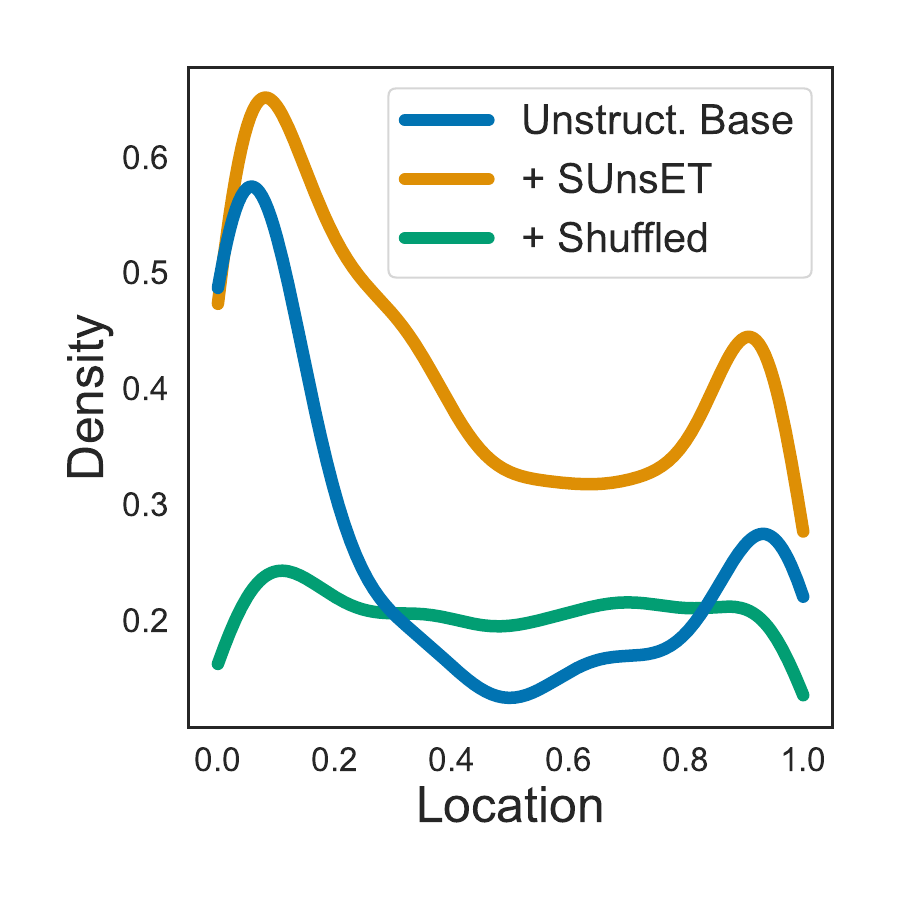}
    \caption{Llama 3.2 3B}
    \end{subfigure}
    ~
    \begin{subfigure}[t]{0.27\textwidth}
        \centering\includegraphics[width=0.88\linewidth]{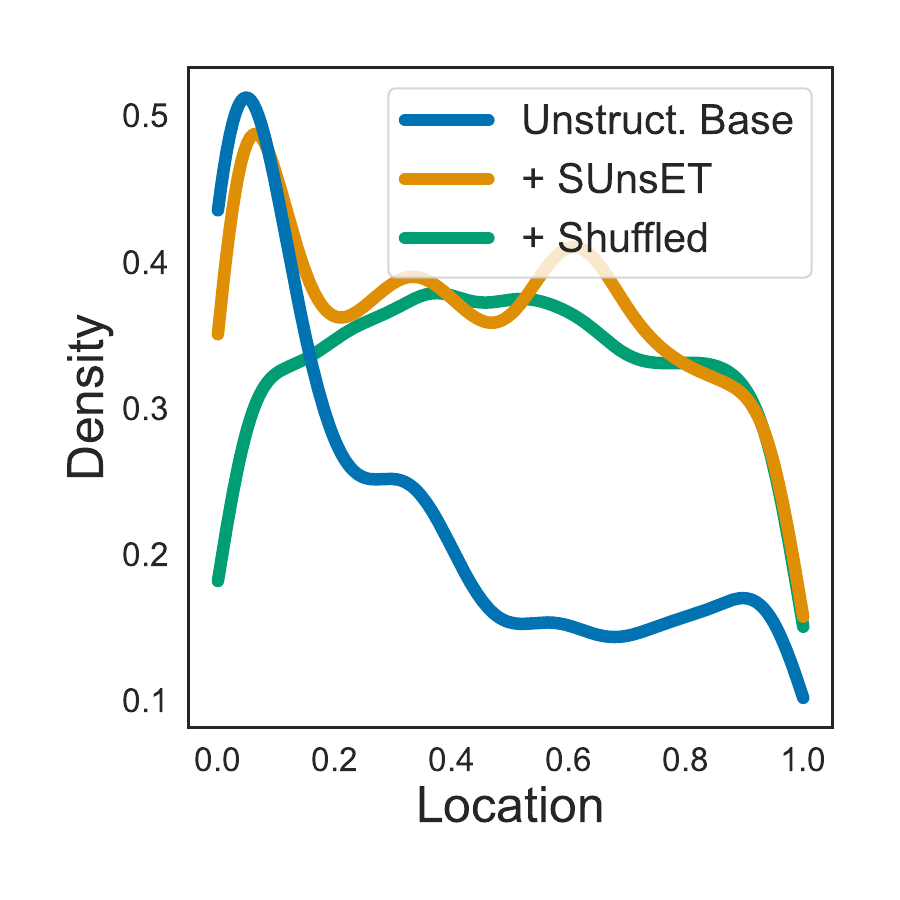}
        \caption{Llama 3.1 8B}
    \end{subfigure}
    ~
    \begin{subfigure}[t]{0.27\textwidth}
        \centering\includegraphics[width=0.88\linewidth]{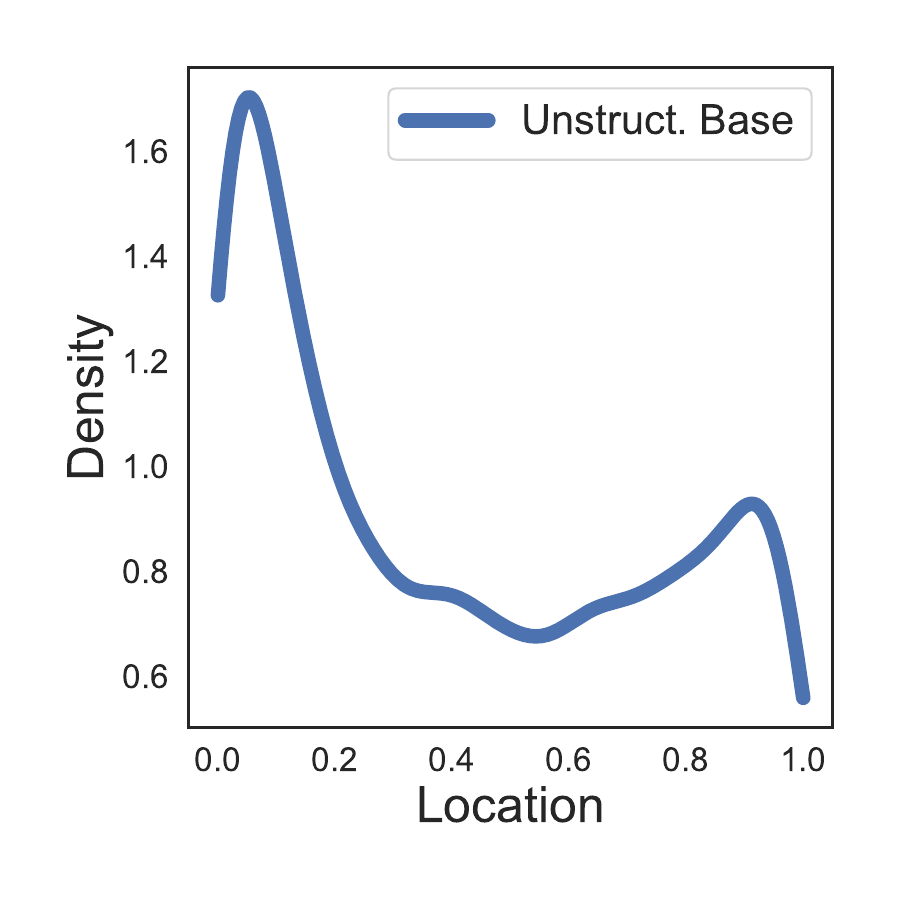}
        \caption{GPT 4o Mini}
    \end{subfigure}
    \\
    \vspace{-3pt}
    \begin{subfigure}[t]{0.27\textwidth}
        \centering\includegraphics[width=0.88\linewidth]{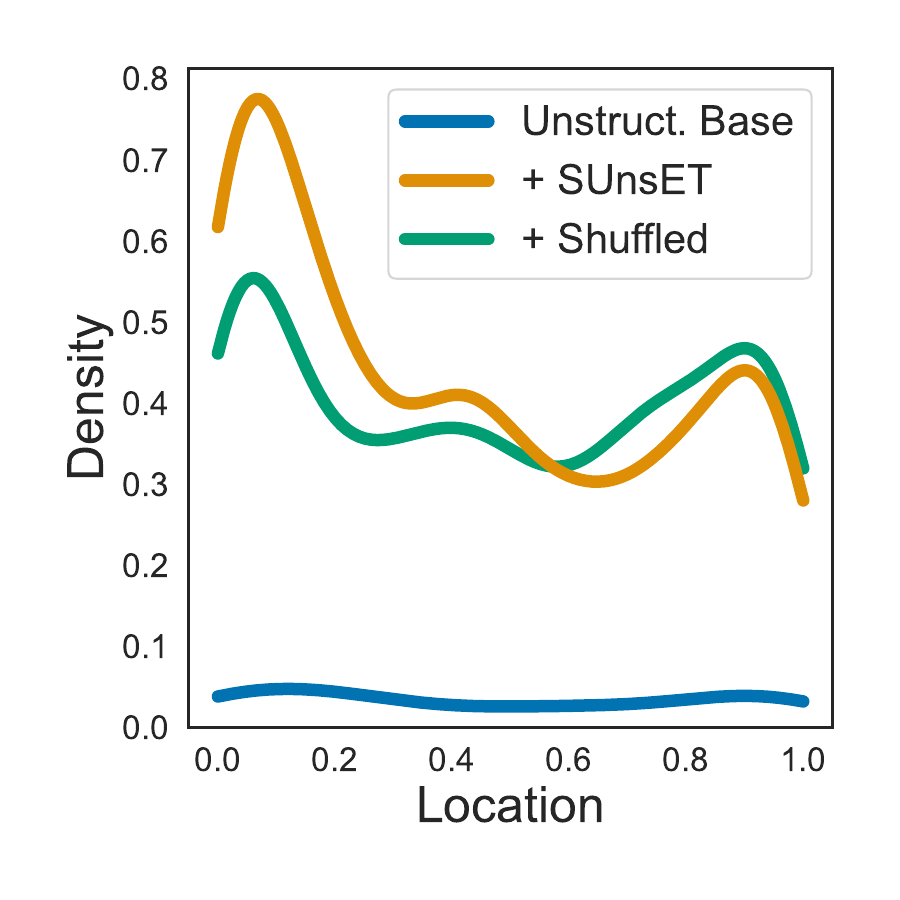}
        \caption{Mistral Nemo 2407}
    \end{subfigure}
    ~
    \begin{subfigure}[t]{0.27\textwidth}
        \centering\includegraphics[width=0.88\linewidth]{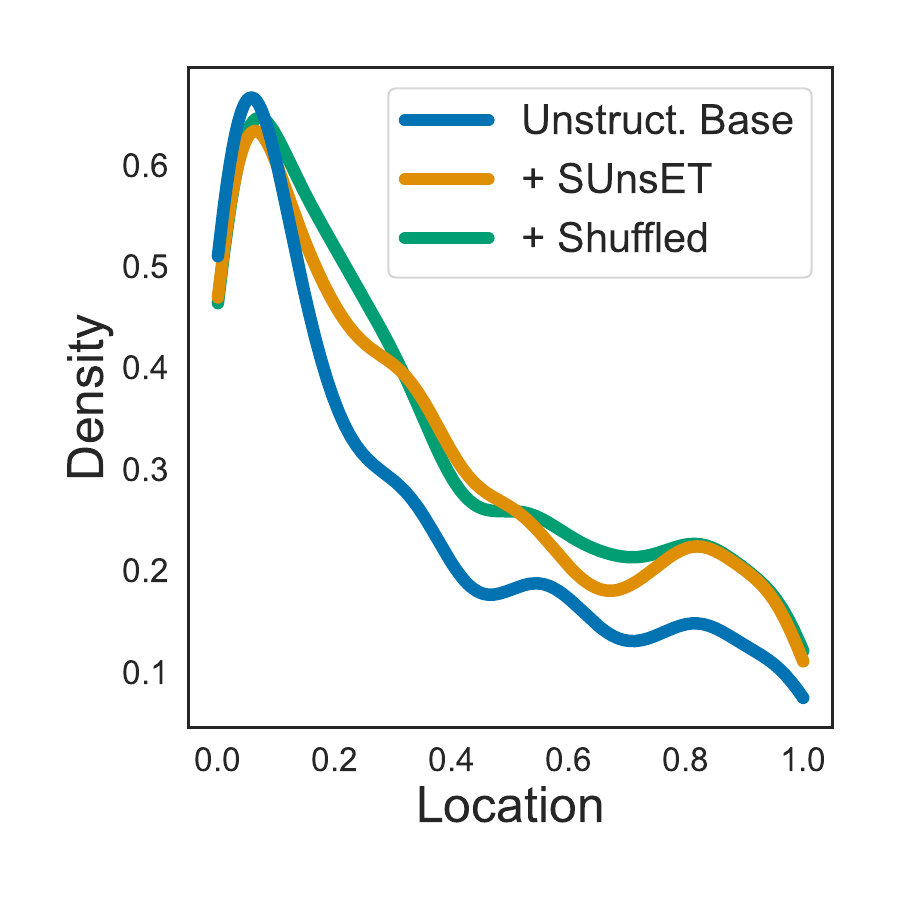}
        \caption{Mixtral 8x7B}
    \end{subfigure}
    ~
    \begin{subfigure}[t]{0.27\textwidth}
        \centering\includegraphics[width=0.88\linewidth]{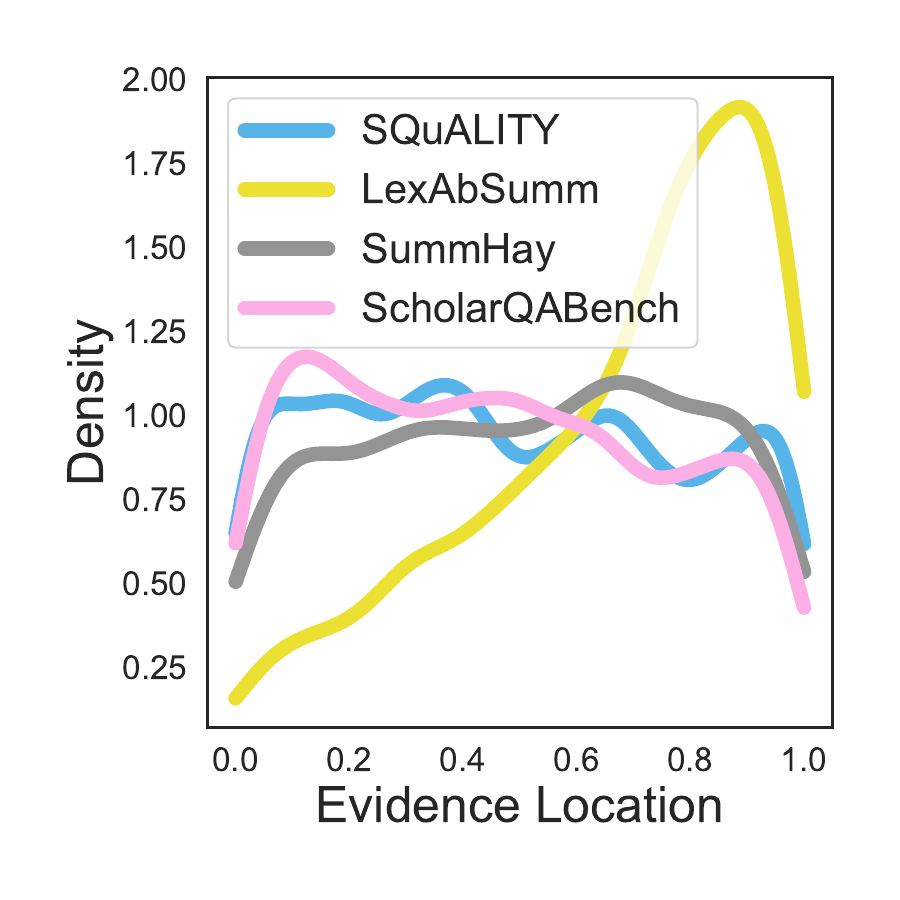}
        \caption{Test Datasets}
    \end{subfigure}
    
\caption{Distribution of location of extracted evidence in the provided source context for different methods. Test dataset evidence location is measured by comparing to reference summaries.}
\label{fig:evidence_location}
\end{figure*}%
    
        
        
        
\begin{figure*}[t]
        \centering
        \includegraphics[width=0.98\linewidth]{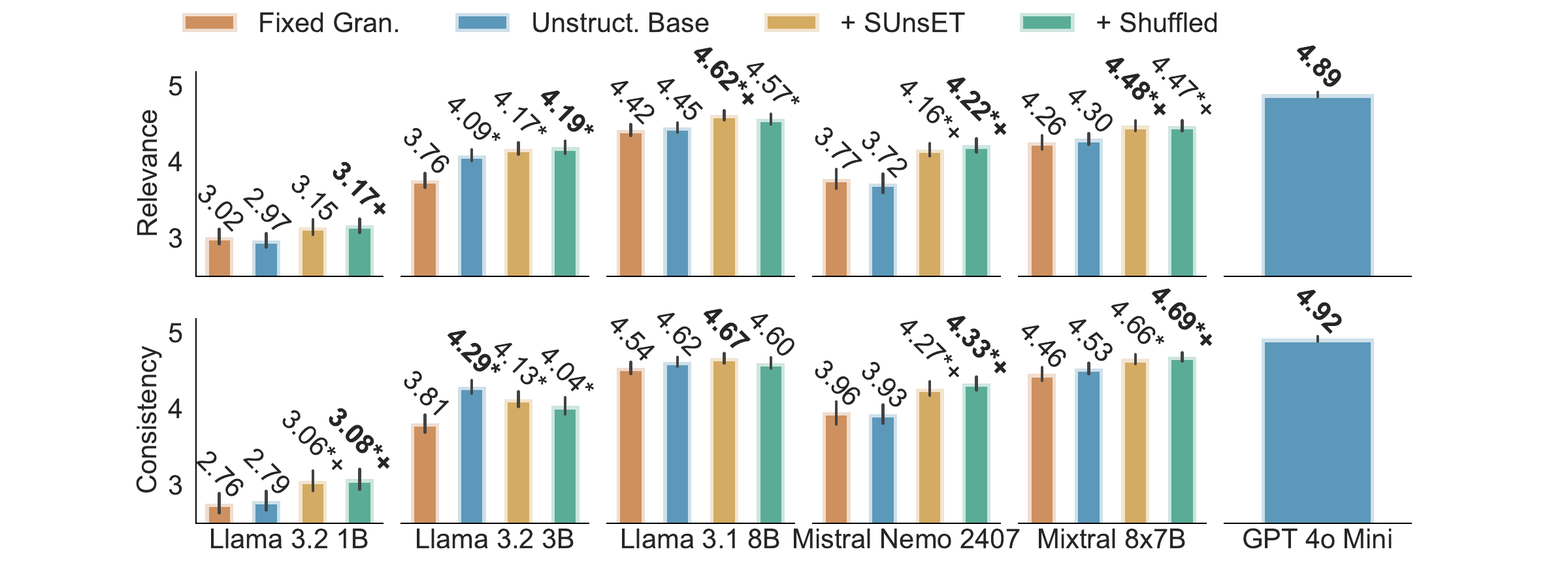}
        \caption{Relevance and consistency of generated summaries.
    \textbf{Bold} best performance for a given model; ``*'' and ``+'' indicate statistical significance above the fixed granularity and non-fine-tuned unstructured baselines, respectively, based on non-overlapping 95\% confidence intervals.}
        \label{fig:abstractive_results}
\end{figure*}

\subsection{RQ1: Can LLMs Use Unstructured Evidence?}

Using the datasets and models just described, we first test how well models can copy and utilize unstructued evidence (i.e., any span of arbitrary length from the context). We look at two aspects: evidence copy accuracy, and evidence quality.

\paragraph{Copy Accuracy} To study copy accuracy, we match each piece of evidence to its longest common substring (LCS) in the context. We present the rate of exact evidence match and 50\% LCS overlap for all models aggregated across all datasets in \autoref{tab:hallucination_results}. We see that \textbf{without fine-tuning, models struggle to copy evidence from the context}. This includes GPT 4o mini, which only copies perfectly 11\% of the time. \textbf{\dataset{} helps models learn to copy evidence spans} in all cases except for the smallest model (Llama 3.2 1B). We see that the number of citations also dramatically increases. 

\paragraph{Evidence Quality} Next, we measure evidence quality based on the relevance and consistency of evidence spans with their citing sentences using the autorater setup previously mentioned. We look at two aspects: the average citation quality (\autoref{fig:citation_avg}) and the citation F1 score (\autoref{fig:citation_results}), which balances citation quality with the total number of sentences that contain a citation. We calculate the latter similarly to \citet{openscholar}: for a given $\langle$summary, evidence\_list$\rangle$ pair, we extract all citations from each sentence and normalize their relevance and consistency scores to lie between $0$ and $100$. For precision, we average these scores over the number of citations, and for recall, we average the scores over the number of sentences in the summary. 

We find that \textbf{the average citation quality of unstructured evidence is better than fixed granularity evidence} (\autoref{fig:citation_avg}). This validates the unstructured evidence approach, where flexible evidence extraction enables higher quality citations to source texts. We also see that models' ability to extract quality evidence is improved by \dataset{}, where our results are on par with GPT 4o Mini. When balancing citation quality and citation quantity (\autoref{fig:citation_results}), we see that \textbf{learning to use unstructured evidence with \dataset{} leads to statistically significant improvements over fixed-granularity and non-fine-tuned baselines across models.} This is particularly the case for medium to larger models. For smaller models (particularly, Llama 3.1 1B), simply fine-tuning for such a complex task is insufficient, where all settings struggle to extract and use evidence. Non-shuffled training is often better than shuffled training, though shuffled training also improves citation quality by a large margin. When balancing for recall, fixed-granularity evidence tends to be better than unstructured evidence without fine-tuning, which makes sense as a model only needs to generate \textit{references} in the fixed-granularity case. Thus, the primary benefits to citation quality by learning from \dataset{} are two-fold: the quality of the evidence itself improves, and the rate of citation improves.

\subsection{RQ2: Is evidence lost-in-the-middle?}

Next, we quantify to what extent unstructured evidence is lost in the middle. For this, we match extracted evidence to its relative location in the document context (based on 50\% LCS overlap) and plot the distributions in \autoref{fig:evidence_location}. As a point of reference, we also plot the distribution of summary sentence locations within the test set documents by matching ground truth reference summaries to their relative locations in their context documents.\footnote{We find the relative location using cosine similarity of S-BERT sentence embeddings~\cite{reimers-2019-sentence-bert}}

We find that \textbf{evidence is lost in the middle for all non-fine-tuned models, most often appearing at the beginning or end of the context}. This includes GPT 4o Mini, which has a sharp spike of evidence in the early context. This stands in contrast to ground truth summary location distributions, which are uniform in all cases except for \las{} which has a bias for evidence at the end of the context. In general, training on \dataset{} without shuffling increases the rate of evidence extraction, and can help decrease the bias. Shuffling on the other hand, increases the rate of evidence extraction and decreases the bias in all cases except for Mixtral 8x7B. Thus, the modular nature of \dataset{} documents, where global structure can be shuffled while local structure is maintained, can be utilized to help reduce positional biases in evidence selection, better reflecting the natural distribution of evidence based on reference data.

\subsection{RQ3: Is Summary Quality Improved?}
Finally, we test if using unstructured evidence has a positive impact on summary quality. To do so, we measure the relevance and consistency of every summary with respect to its context and query. Our results are presented in \autoref{fig:abstractive_results} (results on individual datasets are given in Appendix \ref{sec:individual_results}).

First, \textbf{for fixed granularity evidence the summaries tend to be similar or slightly lower in quality than unstructured with no fine-tuning, further motivating the unstructured approach.} This is likely because the unstructured evidence task has two subtasks: salient evidence selection, followed by summarization, which has been linked to improvements in summary quality~\cite{DBLP:conf/acl/ErnstSSABG0D24}. Second, we find that \textbf{training on \dataset{} leads to statistically significant improvements in summary quality over both baselines.}
Standard and shuffled training on \dataset{} generally lead to similar gains in performance over unstructured with no fine-tuning, meaning the selection of which approach comes down to a tradeoff between overall evidence quality (where standard has a slight edge) and evidence diversity (where shuffled has an edge). To observe the effect of number of training samples from \dataset{}, we perform an ablation 
where we fine-tune on different number of samples in Appendix \ref{sec:training_data_amount} \autoref{fig:variable_data_squality} and \autoref{fig:variable_data_ScholarQABench}, finding that best performance only requires around 3k samples. Third, \textbf{by measuring Pearson's R correlation between citation and summary scores, we find a moderate correlation (0.35 for Relevance and 0.34 for Consistency), demonstrating a relationship between the quality of the citations and the quality of the summaries}. Ultimately, we show the unstructured evidence setup can lead to better evidence and summaries, and demonstrate the utility of \dataset{} for learning the task across diverse, human written data.

\section{Discussion and Conclusion}
Citing precise evidence spans of any arbitrary length for LCQFS has the potential to improve user trust in LLM summaries, as well as the quality of the evidence. Our study highlights salient challenges in this task, contrasts it with the fixed-granular approach, and demonstrates an effective method towards solving it. With no intervention, evidence is lost-in-the-middle, which we show across many settings for the case of unstructured evidence. They additionally struggle to accurately copy arbitrary length evidence from their contexts by default. Our proposed dataset, \dataset{}, serves as a useful and inexpensive synthetic dataset to mitigate these issues. This intervention is at training time, meaning the inference cost is lower than for complex reasoning and inference chains. In addition to improving evidence quality, overall summary quality is improved. We hope this work can be built upon to help create more reliable, trustworthy, and useful summarization systems.

\section*{Acknowledgements}
DW is supported by a Danish Data Science Academy postdoctoral fellowship (grant: 2023-1425). LW is supported in part by the National Science Foundation through grant IIS-2046016. This research was co-funded by Pioneer Centre for AI, DNRF grant number P1.

\section*{Limitations}
 While our approach offers several benefits, there are notable areas to improve upon. Generating unstructured evidence directly can be prone to hallucination, while it is critical for the evidence to be exactly correct. A more precise RAG approach may offer some benefits. While shuffling during training helps the model to pull evidence more evenly, this also reduces the benefits in terms of evidence quality. A more targeted approach based on directly altering positional embeddings may be more appropriate for this \cite{DBLP:conf/acl/HsiehCL0LKGRLKP24}. We experiment with documents using a fixed number of sections in this study; allowing for variable-length documents could deliver greater improvements in performance. Additionally, we acknowledge potential prompt bias influencing model outputs, and that synthetic data may have characteristics which differ from human-written texts. Despite our efforts to mitigate these effects, they persist as a challenge, and using techniques such as APO~\cite{DBLP:conf/emnlp/PryzantI0L0023} could address these issues. Finally, while \dataset{} data is domain agnostic, it could be worth exploring how domain-aware data could help for more targeted applications (e.g., in the legal domain).

 \section*{Ethical Implications}
 LLMs are capable of generating convincing summaries from long contexts, and learning to generate unstructured supporting evidence from the source context can help improve their reliability and transparency. This approach is more flexible than the fixed-granularity approach, but generation will likely always be prone to errors. Validating that generated evidence is authentic is then crucial, as an incorrect citation presented as a ground truth fact could potentially be more harmful than no citation at all. 

 Additionally, synthetic data is clearly useful for learning to cite unstructured evidence. But synthetic data comes with its own ethical issues, including plagiarism and copyright infringement. More work on LLM trust and safety is needed to effectively mitigate this, as we are benefitting technologically from unknowing people's free labor. 

\bibliography{custom}

\appendix
\section{List of Prompts}
\label{sec:prompts}

The full set of prompts used in this study are listed in the figures below.

\subsection{Synthetic Data Generation Prompts}

The prompts used to generated synthetic data are given in \autoref{fig:synth_title_prompt} -- \autoref{fig:validation_prompt}.

\begin{figure*}[t]
    \centering
    \begin{sharp_box}
        \textbf{P1: Title Generation}\\
Imagine that you must write a book. This book can be either fiction or non-fiction.

You can select any subject to write your book about. Please make the book interesting.

Please write a list of 100 possible book titles. 

Please only generate the title for each book. 

Please include a mix of fiction and non-fiction, and please try to cover as many genres as possible.

Please make each book title unique.

Please make the style of each book title as different as possible, and don't repeat title styles.

Please generate titles for books which will have a broad range of appeal. 

Please generate titles for books which will require a broad range of reading levels.

Please try to make each title as different as possible.

Please do not include many titles with a colon (:).

\texttt{\{prev\_titles\_prompt\}}\\

**OUTPUT FORMAT**\\

Please separate each book title with a newline character (``\textbackslash n'')
    \end{sharp_box}
    \caption{Title generation prompt. \texttt{\{prev\_titles\_prompt\}} is filled with prompts of previously generated titles.}
    \label{fig:synth_title_prompt}
\end{figure*}

\begin{figure*}[t]
    \centering
    \begin{sharp_box}
        \textbf{P2: Outline}\\
Imagine that you must write a book. This book can be either fiction or non-fiction.\\

This is the title of your book: \texttt{\{title\}}\\

Please write an outline of this book. Please include the title of the book, and a list of chapters or sections that the book will contain. The book should have 6 sections or chapters.\\

**OUTPUT FORMAT**\\

Please output the outline as a JSON object where the keys are the chapters and the values are a brief outline of the chapter.\\ 

In other words, as:\\

\textasciigrave \textasciigrave \textasciigrave python\\
\{
`Chapter 1': `Chapter 1 outline',\\
`Chapter 2': `Chapter 2 outline',\\
...\\
`Chapter N': `Chapter N outline'\\
\}
\textasciigrave \textasciigrave \textasciigrave
    \end{sharp_box}
    \caption{Outline generation prompt. The \texttt{\{title\}} field is replaced with the title of one document.}
    \label{fig:synth_outline_prompt}
\end{figure*}

\begin{figure*}[t]
    \centering
    \begin{sharp_box}
    \textbf{P3.1: Queries Prompt}\\
Imagine that you must write a book. You are given the following outline of the book\\

\texttt{\{outline\}}\\

Please write a list of 5 questions about the book which summarize the book.\\

Please try to cover different general aspects of the content.\\

Please make the questions very concise.\\

**OUTPUT FORMAT**\\

Please separate each question with a single newline character (``\textbackslash n'')
    \end{sharp_box}
    \caption{Query generation prompt. The \texttt{\{outline\}} is filled with the outline generated by \autoref{fig:synth_outline_prompt}.}
    \label{fig:query_prompt_1}
\end{figure*}

\begin{figure*}[t]
    \centering
    \begin{sharp_box}
    \textbf{P3.2: Initial Summaries and Evidence}\\
Imagine that you are writing a book. This is an outline of the book\\

\texttt{\{outline\}}\\

Please address the following question about the book:\\

\texttt{\{question\}}\\

Please write a summary which addresses the question. Please make the summary as specific and detail oriented as possible. Please include actual examples from the book when possible. Please do not write more than is absolutely necessary.\\

After you write the summary, please write exact quotes and passages you will include in the book, from which the summary could be written. Please include at least
\texttt{\{n\_evidence\}} of these passages, which you intend to include verbatim in the book. Please indicate the exact chapter where the passages will be written in a separate field.\\

**OUTPUT FORMAT**\\

Please a JSON object with two fields: ``summary'', ``evidence'', and ``chapter''. The summary field should have the summary. The evidence field should have a list of evidence sentences from the book. The chapter field should have the exact chapter where the corresponding evidence sentence will appear. Please only indicate the chapter number for this field. There should be the same number of elements in the ``evidence'' field as there are in the ``chapter'' field. In other words, as:\\

\textasciigrave \textasciigrave \textasciigrave python\\
\{\\
`summary': `Summary text',\\
`evidence': $[$`evidence sentence 1', `evidence sentence 2', ...$]$\\
`chapter': $[$1, 4, ...$]$\\
\}\\
\textasciigrave \textasciigrave \textasciigrave
 
    \end{sharp_box}
    \caption{Initial summary and evidence generation prompt. The \texttt{\{outline\}} and \texttt{\{question\}} fields are filled by the output of the previous prompts, while the \texttt{\{n\_evidence\}} field is filled by a random number between 5 and 10.}
    \label{fig:summ_ev_prompt_1}
\end{figure*}

\begin{figure*}[t]
    \centering
    \begin{sharp_box}
    \textbf{P4.1: Document Section Generation}\\
Imagine that you must write a book. You are given the following outline of the book\\

\texttt{\{outline\}}\\

Please write the following chapter of the book in its entirety:\\

\texttt{\{chapter\}}\\

Please also include the following sentences somewhere in the chapter. You must include these passages verbatim (i.e., EXACTLY as is). It is imperative that you do this, otherwise the book will be incomplete:\\

\texttt{\{evidence\}}\\

**OUTPUT FORMAT**\\

Please wrap the content of the chapter you write in a markdown codeblock, in other words, like:\\

\textasciigrave \textasciigrave \textasciigrave \\
content\\
\textasciigrave \textasciigrave \textasciigrave \\
    \end{sharp_box}
    \caption{Document section generation prompt. The \texttt{\{chapter\}} field is filled by the title of the section being generated, as given in the outline.}
    \label{fig:document_section_prompt_1}
\end{figure*}

\begin{figure*}[t]
    \centering
    \begin{sharp_box}
    \textbf{P4.2: Evidence Retrieval Prompt}\\
Please read the following book chapter:\\

\texttt{\{chapter\}}\\

The following passage should have been included in the chapter but was not:\\

\texttt{\{passage\}}\\

Please retrieve the passage from the chapter which is CLOSEST to the given passage.\\

**OUTPUT FORMAT**\\

Please wrap the passage in a markdown codeblock, in other words, like:\\

\textasciigrave \textasciigrave \textasciigrave \\
passage\\
\textasciigrave \textasciigrave \textasciigrave
    \end{sharp_box}
    \caption{Prompt to retrieve evidence from the document when previously generated evidence is not included verbatim. The \texttt{\{passage\}} field is filled with one piece of evidence that was supposed to be included in the section.}
    \label{fig:evidence_retrieval_prompt}
\end{figure*}

\begin{figure*}[t]
    \centering
    \begin{sharp_box}
    \textbf{P5.1: Refinement Prompt}\\
Imagine that you are giving an exam about a book. This is the book\\

\texttt{\{book\}}\\

On an exam, you are asked to summarize the book with respect to this question:\\

\texttt{\{question\}}\\

This is the summary that you are grading:\\

\texttt{\{summary\}}\\

Please rewrite this response so that it is totally accurate and fully addresses the question.\\

Please make the response as specific and detail oriented as possible. The following passages from the document should help in crafting the response:\\

\texttt{\{passages\}}\\

**OUTPUT FORMAT**\\

Please wrap the content of the summary you write in a markdown codeblock, in other words, like:\\

\textasciigrave \textasciigrave \textasciigrave \\
content\\
\textasciigrave \textasciigrave \textasciigrave
    \end{sharp_box}
    \caption{Summary refinement prompt after content has been generated. The \texttt{\{book\}} field is filled with the entire document, where each section is concatenated together. Other fields are filled with the output from the previous prompts.}
    \label{fig:refinement_prompt}
\end{figure*}

\begin{figure*}[t]
    \centering
    \begin{sharp_box}
    \textbf{P5.2: Citance generation}\\
Imagine that you have written a research essay about a book. You have also extracted passages from the book which you used to write the essay.\\

Your job is to add citations to the essay which properly reference the passages that you have extracted.\\

Here is the essay:\\

\texttt{\{essay\}}\\

And here are the evidence passages from the book, each of which is given a number:\\

\texttt{\{evidence\}}\\

Please add citations to all citation-worthy statements in the essay using the numbered evidence list, by indicating the citation numbers of the corresponding evidence. 
More specifically, add the citation number at the end of each relevant sentence in the essay before the punctuation mark e.g., `This work shows the effectiveness of problem X $[$1$]$.' when the passage $[$1$]$ in the evidence list provides full support for the statement. 
Only add a citation if it is fully relevant and unambiguously supportive of that sentence. Not all evidences may be relevant, so only cite those that directly support the statement. 
Please do not add any explanations or justifications for the evidence, simply indicate the evidence numbers if they are relevant. 
If a sentence does not use any of the provided evidence, please simply copy the sentence as is and do not add anything to the end of it. 
If multiple evidences support a statement, please cite them together (e.g., $[$1$]$$[$2$]$). 
For each citation-worthy statement, you only need to add at least one citation, so if multiple evidences support the statement, just add the most relevant citation to the sentence.\\

    \end{sharp_box}
    \caption{Prompt to add citation references to sentences based on extracted evidence. The \texttt{\{essay\}} field is filled with a summary and the \texttt{\{evidence\}} field is filled with its corresponding evidence.}
    \label{fig:citance_prompt}
\end{figure*}

\begin{figure*}[t]
    \centering
    \begin{sharp_box}
    \textbf{P6: Validation Prompt}\\
Imagine that you are judging the quality of a summary of a book. This is the book\\

\texttt{\{book\}}\\

Here is a question about the book:\\

\texttt{\{question\}}\\

And here is the summary which addresses the question:\\

\texttt{\{summary\}}\\

Please judge if you think that the summary meets ALL of the following criteria:\\

1) The summary is absolutely faithful to the book (in other words, all of the information in the summary is contained in the book)\\

2) The summary FULLY addresses the question\\

Please think carefully about your answer. If you think that ALL of the criteria are met, please simply respond with ``YES''.\\ 

Otherwise, please simply respond with ``NO''.
    \end{sharp_box}
    \caption{Prompt to add citation references to sentences based on extracted evidence. Fields are filled with the output of previous prompts.}
    \label{fig:validation_prompt}
\end{figure*}

\begin{figure*}[t]
    \centering
    \begin{sharp_box}
    \textbf{Baseline Non-Pipelined Prompt}\\
Imagine that you must write a book. This book can be either fiction or non-fiction.\\

You can select any subject to write your book about. Please make the book interesting.\\

Please perform the following tasks and output everything in as a JSON object:\\

Please write the title of the book. \\
\texttt{\{title\_prompt\}}\\

Then, please write an outline of this book. Please include a list of chapters or sections that the book will contain. The book should have 6 sections or chapters.\\

Then, please write a list of 5 questions about the book which summarize the book.\\

Then, please write a summary for each question which addresses the question.\\

Then, please write the entire contents of the book. The book should be long, and you should write out the ENTIRE content.\\

Then, extract specific passages from the book for each summary which serve as evidence for the summary.\\

**OUTPUT FORMAT**\\
Please create a well-formatted JSON object with the following fields:\\

title: The title of the book (formatted as a string)\\
outline: The outline of the book (formatted as a string)\\
questions: The questions about the book (formated as a list)\\
summaries: The summaries addressing each question (formatted as a list of the same length as ``questions'')\\
document: The full book (formatted as a string)\\
evidence: A list of evidence passages (formatted as a list of the same length as ``questions'')\\
    \end{sharp_box}
    \caption{Baseline non-pipelined prompt that we use as a point of comparison. The field \texttt{\{title\_prompt\}} is empty for the baseline without diversity enforced, and filled with a list of previous titles and the prompt ``Please do not use any of the following titles:''.}
    \label{fig:baseline_prompt}
\end{figure*}

\subsection{Training and Inference Prompt}
The prompt used for training and inference is given in \autoref{fig:generation_prompt}

\subsection{Evaluation Prompts}
The prompt used to measure relevance is given in \autoref{fig:relevance_prompt} and the prompt used to measure consistency is given in \autoref{fig:consistency_prompt}.

\begin{figure*}[t]
    \centering
    \begin{sharp_box}
        \textbf{Training and Inference Prompt}\\
        
        Your task is to read a document and then write an essay which addresses the following question: \texttt{\{question\_text\}} \\

To write your essay, you should read the document and identify key passages which will help guide your response. Extract every passage which is directly relevant for your essay. Please copy each extracted passage to a list in the format specified below. Please copy the exact text of each passage (do NOT paraphrase!). Then, write your essay which addresses the query. 
\\

Please add citations to all citation-worthy statements using the extracted evidence, by indicating the citation numbers of the corresponding evidence. More specifically, add the citation number at the end of each relevant sentence before the punctuation mark e.g., `This work shows the effectiveness of problem X [1].' when the passage [1] in the evidence list provides full support for the statement. Only add a citation if it is fully relevant and unambiguously supportive of that sentence. Not all evidences may be relevant, so only cite those that directly support the statement. Please do not add any explanations or justifications for the evidence, simply indicate the evidence numbers if they are relevant. If a sentence does not use any of the provided evidence, please simply copy the sentence as is and do not add anything to the end of it. If multiple evidences support a statement, please cite them together (e.g., [1][2]). For each citation-worthy statement, you only need to add at least one citation, so if multiple evidences support the statement, just add the most relevant citation to the sentence.
\\

Please limit to only 10 pieces of evidence.
\\

Here is the document: \texttt{\{context\}}
\\

**OUTPUT FORMAT**\\
Output your response as:\\
EVIDENCE:\\
$[$1$]$ Extracted passage 1\\
$[$2$]$ Extracted passage 2\\
...\\
$[$N$]$ Extracted passage N\\
RESPONSE:\\
response\\
    \end{sharp_box}
    \caption{Full prompt used for fine-tuning and inference. The \texttt{\{question\_text\}} field is filled with a single query, and the \texttt{\{context\}} field is filled with the document context.}
    \label{fig:generation_prompt}
\end{figure*}

\begin{figure*}
    \centering
    \begin{sharp_box}
    \textbf{Summary Combination Prompt}\\
Here is a list of summaries of different sections of a document with respect to the query ``\texttt{\{question\_text\}}'':\\

\texttt{\{context\}}\\

Please combine these summaries into a single summary which addresses the query. If a summary mentions that the query is not addressed, please ignore that summary. Please keep all relevant citations in the final summary. Here is a list of the original citations:\\

\texttt{\{evidence\}}\\
    \end{sharp_box}
    \caption{Prompt to combine section summaries into one final summary.}
\end{figure*}

\begin{figure*}[t]
    \centering
    \begin{sharp_box}
        \textbf{Relevance Prompt}\\
        You will be given one summary written for a document based on a query about that document.\\

Your task is to rate the summary on one metric with respect to the query.\\

Please make sure you read and understand these instructions carefully. Please keep this document open while reviewing, and refer to it as needed.\\

Evaluation Criteria: Relevance (1-5) - selection of important content from the source. The summary should include only important information from the source document which is relevant for the query. Annotators were instructed to penalize summaries which contained redundancies, excess information, and information which does not address the query.\\

Evaluation Steps:\\

1. Read the query, the summary, and the source document carefully.\\
2. Compare the summary to the query and the source document and identify the main point of the document which is relevant to the query.\\
3. Assess how well the summary covers the main points of the source document which are relevant to the query, and how much irrelevant or redundant information it contains.\\
4. Assign a relevance score from 1 to 5.\\

Example:\\

Source Text:\\

\texttt{\{document\}}\\

Query:\\

\texttt{\{query\}}\\

Summary:\\

\texttt{\{summary\}}\\

Evaluation Form (scores ONLY): - \texttt{\{Relevance\}}
    \end{sharp_box}
    \caption{Relevance evaluation prompt from \cite{DBLP:journals/corr/abs-2410-23463}. The \texttt{\{document\}} field is filled with the document context and the \texttt{\{summary\}} field is filled with a summary. When used to evaluate summarization, the \texttt{\{query\}} field is filled with the query used to generate the summary. For citation evaluation, the \texttt{\{query\}} field and all references to queries are removed from the prompt.}
    \label{fig:relevance_prompt}
\end{figure*}

\begin{figure*}[t]
    \centering
    \begin{sharp_box}
        \textbf{Consistency Prompt}\\
        You will be given one summary written for a document based on a query about that document.\\

Your task is to rate the summary on one metric.\\

Please make sure you read and understand these instructions carefully. Please keep this document open while reviewing, and refer to it as needed.\\

Evaluation Criteria:\\

Consistency (1-5) - the factual alignment between the summary and the summarized source with respect to the query. A factually consistent summary contains only statements that are entailed by the source document. Annotators were also asked to penalize summaries that contained hallucinated facts.\\

Evaluation Steps:\\

1. Read the source document carefully and identify the main facts and details it presents with respect to the query.\\
2. Read the summary and compare it to the source document. Check if the summary contains any factual errors that are not supported by the source document.\\
3. Assign a score for consistency based on the Evaluation Criteria.\\

Example:\\

Source Text:\\

\texttt{\{document\}}\\

Query:\\

\texttt{\{query\}}\\

Summary:\\

\texttt{\{summary\}}\\

Evaluation Form (scores ONLY): - \texttt{\{Consistency\}}
    \end{sharp_box}
    \caption{Consistency evaluation prompt from \cite{DBLP:journals/corr/abs-2410-23463}. The \texttt{\{document\}} field is filled with the document context and the \texttt{\{summary\}} field is filled with a summary. When used to evaluate summarization, the \texttt{\{query\}} field is filled with the query used to generate the summary. For citation evaluation, the \texttt{\{query\}} field and all references to queries are removed from the prompt.}
    \label{fig:consistency_prompt}
\end{figure*}

\section{Full Dataset Descriptions}
\label{sec:datasets_appendix}
The test datasets we use in this study include:
 \paragraph{SQuALITY}~\cite{DBLP:conf/emnlp/WangPCPB22} is a single-document task created from public domain short sci-fi stories where expert annotators create original summaries, providing both an overall narrative and detailed responses to specific questions, challenging models to capture broad context as well as fine-grained information.
 \paragraph{LexAbSumm}~\cite{DBLP:conf/coling/SantoshAG24} is a single-document task which contains legal judgments from the European Court of Human Rights, focusing on aspect-specific summaries that distill complex legal arguments. 
 \paragraph{SummHay}~\cite{DBLP:conf/emnlp/LabanFXW24} is a multi-document task composed of large-scale ``haystacks'' of documents with embedded ``insights'' which are relevant to the queries.
 \paragraph{ScholarQABench}~\cite{openscholar} is a multi-document task focused on scientific literature, comprising expert-crafted queries and extended answers drawn from a broad corpus of open-access research papers.

 \section{Topic Diversity Comparison}
 \label{sec:topic_diversity}

 We have measured the topic diversity of SUnsET using the topic diversity approach from \cite{DBLP:conf/eacl/TerragniFGTC21}. This uses LDA to identify 200 topics across each document, sums up the number of unique words in the first 200 words of each topic, and averages this over a maximum of 200 words * 200 topics (so the score is 1 if each topic has at least 200 unique words, see \url{https://github.com/MIND-Lab/OCTIS}). We compare this to the two baseline datasets, as well as the human test data, finding that the data in SUnsET is indeed diverse and comparable to human data.

\section{Results on Individual Datasets}
\label{sec:individual_results}

Results on individual datasets are given in \autoref{tab:retrieval_results_precision} (citation precision), \autoref{tab:retrieval_results_recall} (citation recall), and \autoref{tab:retrieval_results_F1} (F1 score based on citation precision and recall). We see that citation precision is almost uniformly improved across datasets when using unstructured evidence. In other words, when evidence is used within a summary, the evidence is higher quality than fixed granularity evidence in all but 3 cases. This quality is generally further improved by learning from \dataset{}. Recall is also improved by learning from \dataset{}, and is often better than fixed granularity evidence where a model simply needs to generate reference numbers (as opposed to unstructured where the evidence must also be copied, making the task more challenging). For Llama 3.1 8B and Nemo, overall F1 score is better across all datasets, while for Mixtral and the smaller Llama models the results are mixed across datasets. This is generally because the recall of the fixed granular case tends to be slightly higher, despite referencing lower quality evidences on average. However, when looking at the averages across datasets (\autoref{fig:citation_results}), we see that learning to cite unstructured evidence with \dataset{} leads to the best overall performance.

For summary quality (\autoref{tab:abstractive_results}), unstructured evidence leads to the best summaries across models and datasets most often, including the best overall performance with \dataset{} fine-tuned models within each dataset. The results on smaller models are more mixed across datasets, likely due to the difficulty for smaller models to learn the unstructured evidence task in general. Learning from \dataset{} appears to be especially useful for improving summaries on multi-document datasets (\smh{} and \sqa{}), which always see improvements over the unstructured baseline.

\begin{table*}[t]
    \centering

    \begin{tabular}{l c c | c c | c c | c c}
    \toprule 
    & \multicolumn{2}{ c |}{ SLT\textsuperscript{S} } & \multicolumn{2}{c|}{ LAS\textsuperscript{S} } & \multicolumn{2}{c |}{ SMH\textsuperscript{M} } & \multicolumn{2}{c}{ SQB\textsuperscript{M} }\\
    \midrule
    Model & Rel\textsubscript{Prec} & Con\textsubscript{Prec} & Rel\textsubscript{Prec} & Con\textsubscript{Prec} & Rel\textsubscript{Prec} & Con\textsubscript{Prec} & Rel\textsubscript{Prec} & Con\textsubscript{Prec}\\

\midrule
Llama 3.2 1B &  $12.50$ & $12.50$ & $30.94$ & $20.51$ & $\underline{50.00}$ & $0.00$ & $37.50$ & $\underline{50.00}$ \\
Fixed Gran. & $19.86$ & $4.10$ & $39.22$ & $25.86$ & $25.94$ & $\underline{8.88}$ & $21.82$ & $11.47$ \\
+ SUnsET & $18.80$ & $10.61$ & $41.27$ & $32.05$ & $0.00$ & $0.00$ & $45.18$ & $24.08$ \\
+ Shuffled & $\underline{28.60}$ & $\underline{13.01}$ & $\underline{50.34}$ & $\underline{48.86}$ & $\underline{50.00}$ & $0.00$ & $\underline{62.38}$ & $48.20$ \\\midrule
Llama 3.2 3B &  $34.27$ & $20.34$ & $62.30$ & $55.77$ & $54.34$ & $44.53$ & $52.39$ & $39.86$ \\
Fixed Gran. & $34.84$ & $15.24$ & $62.02$ & $\underline{56.35}$ & $24.59$ & $24.91$ & $35.86$ & $29.97$ \\
+ SUnsET & $\underline{45.17}$ & $25.65$ & $61.16$ & $53.96$ & $64.75$ & $59.25$ & $52.91$ & $45.00$ \\
+ Shuffled & $44.28$ & $\underline{27.20}$ & $\underline{62.76}$ & $54.42$ & $\underline{65.76}$ & $\underline{62.84}$ & $\underline{60.98}$ & $\underline{56.37}$ \\\midrule
Llama 3.1 8B &  $42.69$ & $27.70$ & $67.18$ & $61.79$ & $62.72$ & $57.14$ & $49.95$ & $39.24$ \\
Fixed Gran. & $44.45$ & $26.84$ & $59.66$ & $54.80$ & $39.14$ & $39.00$ & $50.21$ & $49.70$ \\
+ SUnsET & $50.91$ & $33.71$ & $\underline{75.21}$ & $\underline{70.45}$ & $\mathbf{\underline{\mathbf{74.31}}}$ & $\underline{70.96}$ & $\underline{67.36}$ & $\underline{61.17}$ \\
+ Shuffled & $\underline{53.13}$ & $\underline{36.79}$ & $\mathbf{73.78}$ & $68.99$ & $70.55$ & $67.15$ & $64.70$ & $61.12$ \\\midrule
Mistral Nemo 2407 &  $31.67$ & $14.00$ & $60.27$ & $53.41$ & $\underline{\mathbf{73.78}}$ & $\mathbf{\underline{\mathbf{73.78}}}$ & $69.49$ & $61.38$ \\
Fixed Gran. & $32.44$ & $19.12$ & $60.28$ & $54.00$ & $29.59$ & $25.97$ & $37.86$ & $28.03$ \\
+ SUnsET & $\mathbf{\underline{\mathbf{57.34}}}$ & $36.90$ & $78.96$ & $\underline{78.69}$ & $73.62$ & $70.84$ & $\mathbf{\underline{\mathbf{71.44}}}$ & $\underline{66.50}$ \\
+ Shuffled & $56.07$ & $\underline{38.18}$ & $\underline{78.97}$ & $78.39$ & $70.58$ & $65.37$ & $64.97$ & $61.20$ \\\midrule
Mixtral 8x7B &  $47.82$ & $32.79$ & $81.58$ & $83.76$ & $68.54$ & $66.53$ & $53.67$ & $48.02$ \\
Fixed Gran. & $43.78$ & $24.11$ & $64.14$ & $61.01$ & $37.43$ & $29.62$ & $61.32$ & $\mathbf{\underline{\mathbf{67.63}}}$ \\
+ SUnsET & $50.74$ & $35.96$ & $82.94$ & $82.94$ & $69.77$ & $69.82$ & $60.82$ & $57.49$ \\
+ Shuffled & $\underline{52.52}$ & $\mathbf{\underline{\mathbf{38.71}}}$ & $\mathbf{\underline{\mathbf{84.19}}}$ & $\mathbf{\underline{\mathbf{85.29}}}$ & $\underline{73.80}$ & $\underline{73.33}$ & $\underline{61.94}$ & $59.22$ \\\midrule
\midrule GPT 4o Mini &  \textit{60.11} & \textit{52.11} & \textit{77.92} & \textit{74.76} & \textit{77.09} & \textit{75.57} & \textit{57.49} & \textit{49.18} \\
    \bottomrule 

    \end{tabular}
    \caption{Relevance and consistency \textbf{precision} of evidence sentences with respect to their citances.
    Precision measures the average citation quality within a given summary.
    \textbf{Bold} indicates best overall performance, \underline{Underline} indicates best performance for individual models.
    \textsuperscript{S} indicates single document tasks, \textsuperscript{M} indicates multi-document. SQ is \sqa{}, LAS is \las{}, SMH is \smh{}, and SQB is \ops{}}
    \label{tab:retrieval_results_precision}
\end{table*}

\begin{table*}[t]
    \centering

    \begin{tabular}{l c c | c c | c c | c c}
    \toprule 
    & \multicolumn{2}{ c |}{ SLT\textsuperscript{S} } & \multicolumn{2}{c|}{ LAS\textsuperscript{S} } & \multicolumn{2}{c |}{ SMH\textsuperscript{M} } & \multicolumn{2}{c}{ SQB\textsuperscript{M} }\\
    \midrule
    Model & Rel\textsubscript{Rec} & Con\textsubscript{Rec} & Rel\textsubscript{Rec} & Con\textsubscript{Rec} & Rel\textsubscript{Rec} & Con\textsubscript{Rec} & Rel\textsubscript{Rec} & Con\textsubscript{Rec}\\

\midrule
Llama 3.2 1B &  $0.10$ & $0.10$ & $0.94$ & $0.69$ & $0.27$ & $0.00$ & $0.06$ & $0.08$ \\
Fixed Gran. & $0.33$ & $0.12$ & $\underline{5.24}$ & $\underline{3.42}$ & $\underline{0.28}$ & $\underline{0.15}$ & $1.88$ & $\underline{1.14}$ \\
+ SUnsET & $0.82$ & $0.43$ & $4.06$ & $2.45$ & $0.00$ & $0.00$ & $\underline{2.40}$ & $0.93$ \\
+ Shuffled & $\underline{1.26}$ & $\underline{0.52}$ & $2.01$ & $1.94$ & $0.05$ & $0.00$ & $0.48$ & $0.41$ \\\midrule
Llama 3.2 3B &  $4.85$ & $2.82$ & $11.64$ & $10.13$ & $5.75$ & $4.90$ & $11.22$ & $8.36$ \\
Fixed Gran. & $18.13$ & $7.45$ & $\underline{39.63}$ & $\underline{35.85}$ & $0.93$ & $0.78$ & $\underline{24.02}$ & $\underline{20.37}$ \\
+ SUnsET & $\underline{20.14}$ & $\underline{11.86}$ & $26.95$ & $23.70$ & $\underline{26.68}$ & $\underline{24.54}$ & $10.18$ & $8.80$ \\
+ Shuffled & $11.09$ & $6.85$ & $14.56$ & $12.55$ & $22.24$ & $20.82$ & $11.53$ & $11.07$ \\\midrule
Llama 3.1 8B &  $8.90$ & $5.61$ & $22.41$ & $20.76$ & $25.52$ & $23.23$ & $16.68$ & $13.17$ \\
Fixed Gran. & $14.88$ & $8.98$ & $36.83$ & $33.73$ & $12.22$ & $12.19$ & $33.55$ & $\underline{32.60}$ \\
+ SUnsET & $\underline{21.32}$ & $\underline{14.28}$ & $\mathbf{\underline{\mathbf{41.31}}}$ & $\mathbf{\underline{\mathbf{38.72}}}$ & $\mathbf{\underline{\mathbf{47.39}}}$ & $\mathbf{\underline{\mathbf{45.45}}}$ & $\mathbf{\underline{\mathbf{35.28}}}$ & $32.47$ \\
+ Shuffled & $16.80$ & $11.70$ & $35.13$ & $32.78$ & $42.35$ & $40.44$ & $32.31$ & $30.86$ \\\midrule
Mistral Nemo 2407 &  $0.47$ & $0.20$ & $1.13$ & $1.08$ & $5.18$ & $5.17$ & $4.94$ & $4.54$ \\
Fixed Gran. & $5.39$ & $3.26$ & $10.40$ & $9.34$ & $2.64$ & $2.39$ & $12.04$ & $8.79$ \\
+ SUnsET & $\underline{17.48}$ & $\underline{11.30}$ & $\underline{19.93}$ & $\underline{19.66}$ & $\underline{16.63}$ & $\underline{15.80}$ & $\underline{17.68}$ & $\underline{16.59}$ \\
+ Shuffled & $13.81$ & $9.38$ & $19.59$ & $19.14$ & $16.17$ & $15.06$ & $13.54$ & $13.00$ \\\midrule
Mixtral 8x7B &  $15.47$ & $11.04$ & $29.99$ & $30.85$ & $29.87$ & $28.54$ & $13.92$ & $12.46$ \\
Fixed Gran. & $\mathbf{\underline{\mathbf{33.32}}}$ & $\mathbf{\underline{\mathbf{18.68}}}$ & $\underline{36.40}$ & $\underline{34.42}$ & $6.32$ & $5.75$ & $\underline{34.11}$ & $\mathbf{\underline{\mathbf{37.82}}}$ \\
+ SUnsET & $19.06$ & $13.64$ & $30.65$ & $30.68$ & $37.91$ & $37.31$ & $23.06$ & $21.80$ \\
+ Shuffled & $20.40$ & $15.40$ & $31.82$ & $32.08$ & $\underline{39.55}$ & $\underline{38.65}$ & $27.00$ & $26.22$ \\\midrule
\midrule GPT 4o Mini &  \textit{28.38} & \textit{23.86} & \textit{51.15} & \textit{49.07} & \textit{55.03} & \textit{53.93} & \textit{25.82} & \textit{21.99} \\
    \bottomrule 

    \end{tabular}
    \caption{Relevance and consistency \textbf{recall} of evidence sentences with respect to their citances.
    Recall measures citation quality and averages based on the total number of sentences in a summary. This penalizes models which produce fewer citations.
    \textbf{Bold} indicates best overall performance, \underline{Underline} indicates best performance for individual models.
    \textsuperscript{S} indicates single document tasks, \textsuperscript{M} indicates multi-document. SQ is \sqa{}, LAS is \las{}, SMH is \smh{}, and SQB is \ops{}}
    \label{tab:retrieval_results_recall}
\end{table*}

\begin{table*}[t]
    \centering

    \begin{tabular}{l c c | c c | c c | c c}
    \toprule 
    & \multicolumn{2}{ c |}{ SLT\textsuperscript{S} } & \multicolumn{2}{c|}{ LAS\textsuperscript{S} } & \multicolumn{2}{c |}{ SMH\textsuperscript{M} } & \multicolumn{2}{c}{ SQB\textsuperscript{M} }\\
    \midrule
    Model & Rel\textsubscript{F1} & Con\textsubscript{F1} & Rel\textsubscript{F1} & Con\textsubscript{F1} & Rel\textsubscript{F1} & Con\textsubscript{F1} & Rel\textsubscript{F1} & Con\textsubscript{F1}\\

\midrule
Llama 3.2 1B &  $0.14$ & $0.14$ & $1.22$ & $0.84$ & $0.36$ & $0.00$ & $0.11$ & $0.14$ \\
Fixed Gran. & $0.40$ & $0.13$ & $\underline{6.67}$ & $\underline{4.40}$ & $\underline{0.39}$ & $\underline{0.19}$ & $2.27$ & $\underline{1.35}$ \\
+ SUnsET & $1.18$ & $0.62$ & $5.43$ & $3.59$ & $0.00$ & $0.00$ & $\underline{3.13}$ & $1.26$ \\
+ Shuffled & $\underline{1.85}$ & $\underline{0.80}$ & $3.14$ & $3.04$ & $0.08$ & $0.00$ & $0.85$ & $0.72$ \\\midrule
Llama 3.2 3B &  $6.61$ & $3.86$ & $15.17$ & $13.29$ & $7.66$ & $6.52$ & $14.12$ & $10.54$ \\
Fixed Gran. & $21.71$ & $9.02$ & $\underline{45.80}$ & $\underline{41.44}$ & $1.37$ & $1.13$ & $\underline{27.77}$ & $\underline{23.49}$ \\
+ SUnsET & $\underline{25.36}$ & $\underline{14.76}$ & $33.42$ & $29.40$ & $\underline{32.21}$ & $\underline{29.59}$ & $13.76$ & $11.85$ \\
+ Shuffled & $15.14$ & $9.33$ & $19.45$ & $16.80$ & $26.78$ & $25.15$ & $17.45$ & $16.55$ \\\midrule
Llama 3.1 8B &  $11.66$ & $7.38$ & $28.89$ & $26.76$ & $32.07$ & $29.17$ & $20.73$ & $16.32$ \\
Fixed Gran. & $18.90$ & $11.32$ & $42.44$ & $38.86$ & $14.29$ & $14.23$ & $38.56$ & $37.64$ \\
+ SUnsET & $\underline{27.69}$ & $\underline{18.48}$ & $\mathbf{\underline{\mathbf{50.78}}}$ & $\mathbf{\underline{\mathbf{47.62}}}$ & $\mathbf{\underline{\mathbf{53.62}}}$ & $\mathbf{\underline{\mathbf{51.43}}}$ & $\mathbf{\underline{\mathbf{44.03}}}$ & $\underline{40.49}$ \\
+ Shuffled & $23.13$ & $16.12$ & $44.16$ & $41.18$ & $48.72$ & $46.50$ & $41.49$ & $39.59$ \\\midrule
Mistral Nemo 2407 &  $0.53$ & $0.23$ & $1.36$ & $1.29$ & $6.68$ & $6.68$ & $6.08$ & $5.54$ \\
Fixed Gran. & $6.61$ & $3.93$ & $13.36$ & $11.95$ & $3.71$ & $3.36$ & $15.05$ & $11.03$ \\
+ SUnsET & $\underline{21.71}$ & $\underline{13.99}$ & $\underline{23.38}$ & $\underline{23.09}$ & $\underline{20.73}$ & $\underline{19.71}$ & $\underline{22.00}$ & $\underline{20.61}$ \\
+ Shuffled & $17.67$ & $11.96$ & $22.85$ & $22.42$ & $19.82$ & $18.38$ & $16.87$ & $16.14$ \\\midrule
Mixtral 8x7B &  $17.83$ & $12.64$ & $34.27$ & $35.23$ & $33.40$ & $32.02$ & $17.30$ & $15.48$ \\
Fixed Gran. & $\mathbf{\underline{\mathbf{36.35}}}$ & $\mathbf{\underline{\mathbf{20.33}}}$ & $\underline{42.34}$ & $\underline{40.15}$ & $8.45$ & $7.46$ & $\underline{40.06}$ & $\mathbf{\underline{\mathbf{44.4}0}}$ \\
+ SUnsET & $22.60$ & $16.11$ & $35.81$ & $35.81$ & $42.91$ & $42.27$ & $28.61$ & $26.94$ \\
+ Shuffled & $23.79$ & $17.85$ & $37.21$ & $37.57$ & $\underline{43.89}$ & $\underline{42.98}$ & $32.25$ & $31.16$ \\\midrule
\midrule GPT 4o Mini &  \textit{37.39} & \textit{31.70} & \textit{61.17} & \textit{58.68} & \textit{63.61} & \textit{62.35} & \textit{33.71} & \textit{28.63} \\
    \bottomrule 

    \end{tabular}
    \caption{Relevance and consistency \textbf{F1} of evidence sentences with respect to their citances.
    We follow a similar setup to \cite{{DBLP:conf/emnlp/LabanFXW24,openscholar}} where we measure citation precision and recall in order to calculate an overall F1 score for both relevance and consistency.
    \textbf{Bold} indicates best overall performance, \underline{Underline} indicates best performance for individual models.
    \textsuperscript{S} indicates single document tasks, \textsuperscript{M} indicates multi-document. SQ is \sqa{}, LAS is \las{}, SMH is \smh{}, and SQB is \ops{}}
    \label{tab:retrieval_results_F1}
\end{table*}

\begin{table*}[t]
    \centering
    
    \begin{tabular}{l c c | c c | c c | c c }
    \toprule 
    & \multicolumn{2}{ c |}{ SLT\textsuperscript{S} } & \multicolumn{2}{c|}{ LAS\textsuperscript{S} } & \multicolumn{2}{c |}{ SMH\textsuperscript{M} } & \multicolumn{2}{c |}{ SQB\textsuperscript{M} }\\
    \midrule
    Model & Rel & Con & Rel & Con & Rel & Con & Rel & Con\\

\midrule
Llama 3.2 1B &  $2.28$ & $1.63$ & $3.09$ & $\underline{2.88}$ & $3.52$ & $3.70$ & $2.90$ & $2.93$ \\
Fixed Gran. & $2.42$ & $1.49$ & $\underline{3.28}$ & $2.81$ & $3.09$ & $3.32$ & $\underline{3.28}$ & $\underline{3.36}$ \\
+ SUnsET & $\underline{2.60}$ & $\underline{2.23}$ & $2.99$ & $2.75$ & $3.82$ & $4.04$ & $3.17$ & $3.02$ \\
+ Shuffled & $2.57$ & $2.15$ & $3.06$ & $2.78$ & $\underline{3.83}$ & $\underline{4.35}$ & $3.18$ & $3.07$ \\\midrule
Llama 3.2 3B &  $\underline{3.66}$ & $\underline{3.52}$ & $\underline{4.26}$ & $\underline{4.49}$ & $4.47$ & $4.83$ & $3.99$ & $4.21$ \\
Fixed Gran. & $3.40$ & $3.11$ & $4.12$ & $4.34$ & $3.45$ & $3.53$ & $4.04$ & $\underline{4.28}$ \\
+ SUnsET & $3.49$ & $3.10$ & $4.13$ & $4.17$ & $4.73$ & $4.91$ & $4.26$ & $4.20$ \\
+ Shuffled & $3.16$ & $2.68$ & $4.17$ & $4.13$ & $\underline{4.88}$ & $\underline{4.95}$ & $\underline{4.36}$ & $4.20$ \\\midrule
Llama 3.1 8B &  $\underline{4.26}$ & $\underline{4.44}$ & $4.60$ & $\mathbf{\underline{4.81}}$ & $4.84$ & $4.92$ & $4.07$ & $4.24$ \\
Fixed Gran. & $4.23$ & $4.34$ & $4.59$ & $4.79$ & $4.43$ & $4.55$ & $4.52$ & $4.59$ \\
+ SUnsET & $4.23$ & $4.24$ & $4.65$ & $\mathbf{\underline{4.81}}$ & $4.89$ & $\mathbf{\underline{4.98}}$ & $4.58$ & $4.55$ \\
+ Shuffled & $4.08$ & $4.02$ & $\mathbf{\underline{4.66}}$ & $4.75$ & $\mathbf{\underline{4.92}}$ & $\mathbf{\underline{4.98}}$ & $\mathbf{\underline{4.68}}$ & $\mathbf{\underline{4.69}}$ \\\midrule
Mistral Nemo 2407 &  $4.15$ & $4.15$ & $3.52$ & $3.70$ & $4.05$ & $4.37$ & $3.09$ & $3.25$ \\
Fixed Gran. & $4.12$ & $4.26$ & $\underline{4.42}$ & $\underline{4.68}$ & $2.54$ & $2.62$ & $\underline{4.06}$ & $\underline{4.23}$ \\
+ SUnsET & $4.29$ & $4.31$ & $4.24$ & $4.39$ & $\underline{4.52}$ & $4.66$ & $3.65$ & $3.77$ \\
+ Shuffled & $\underline{4.41}$ & $\underline{4.38}$ & $4.35$ & $4.46$ & $4.50$ & $\underline{4.73}$ & $3.76$ & $3.86$ \\\midrule
Mixtral 8x7B &  $4.21$ & $4.47$ & $4.43$ & $4.73$ & $4.46$ & $4.67$ & $4.09$ & $4.27$ \\
Fixed Gran. & $4.46$ & $4.63$ & $4.46$ & $4.71$ & $3.93$ & $4.08$ & $4.19$ & $\underline{4.43}$ \\
+ SUnsET & $4.48$ & $4.64$ & $4.54$ & $4.79$ & $4.49$ & $4.74$ & $\underline{4.29}$ & $\underline{4.43}$ \\
+ Shuffled & $\mathbf{\underline{4.55}}$ & $\mathbf{\underline{4.67}}$ & $\underline{4.56}$ & $\mathbf{\underline{4.81}}$ & $\underline{4.55}$ & $\underline{4.78}$ & $4.20$ & $\underline{4.43}$ \\\midrule
\midrule GPT 4o Mini &  \textit{4.77} & \textit{4.85} & \textit{4.87} & \textit{4.93} & \textit{4.98} & \textit{5.00} & \textit{4.93} & \textit{4.94} \\
    \bottomrule 

    \end{tabular}
    \caption{Relevance and consistency of generated summaries. Relevance and consistency are measured using an autorater (DeepSeek-V3)~\cite{DBLP:conf/emnlp/LiuIXWXZ23} based on previously validated prompts~\cite{DBLP:journals/corr/abs-2410-23463}. 
    \textbf{Bold} indicates best overall performance, \underline{Underline} indicates best performance for individual models.
    \textsuperscript{S} indicates single document tasks, \textsuperscript{M} indicates multi-document. SQ is \sqa{}, LAS is \las{}, SMH is \smh{}, and SQB is \ops{}.}
    \label{tab:abstractive_results}
\end{table*}

\section{Training Data Requirements}
\label{sec:training_data_amount}
\begin{figure*}
    \centering
    \begin{subfigure}[t]{0.32\textwidth}
    \centering
    \includegraphics[width=1\linewidth]{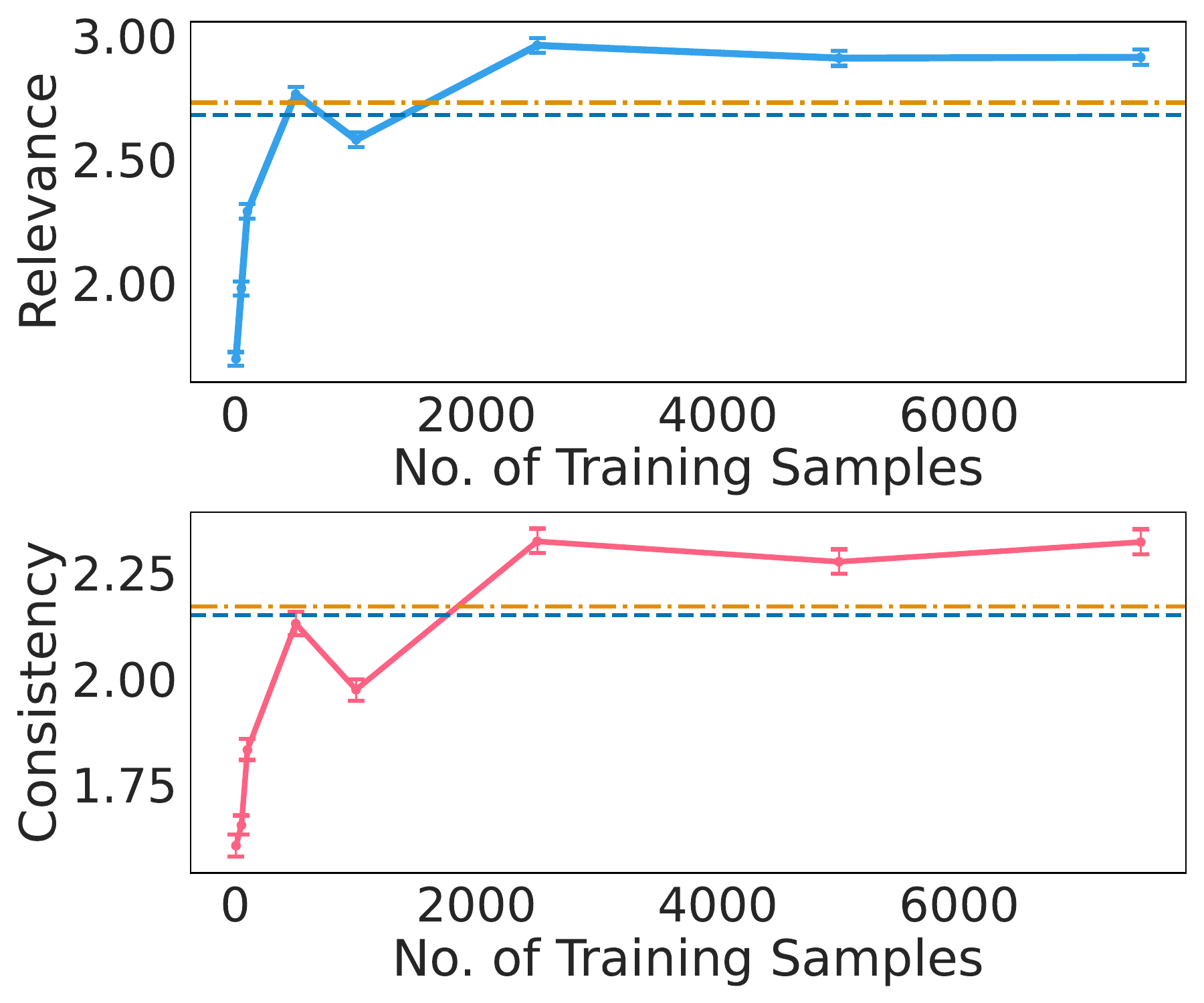}
    \vspace{1pt}
    \caption{Llama 3.2 1B}
    \end{subfigure}
    ~
    \begin{subfigure}[t]{0.32\textwidth}
        \centering\includegraphics[width=1\linewidth]{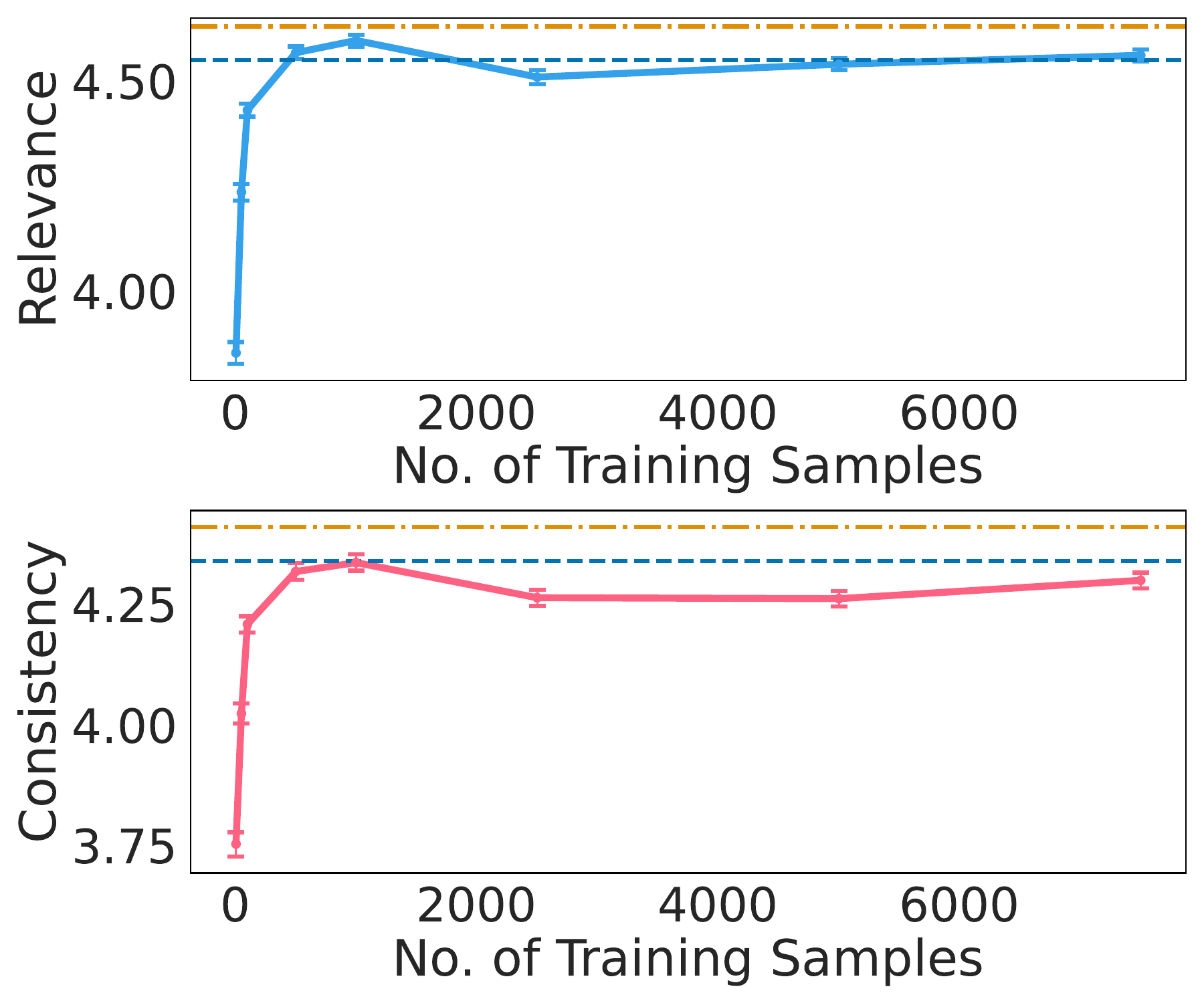}
    \vspace{1pt}
        
    \caption{Llama 3.1 8B}
    \end{subfigure}
    \begin{subfigure}[t]{0.32\textwidth}
        \centering\includegraphics[width=1 \linewidth]{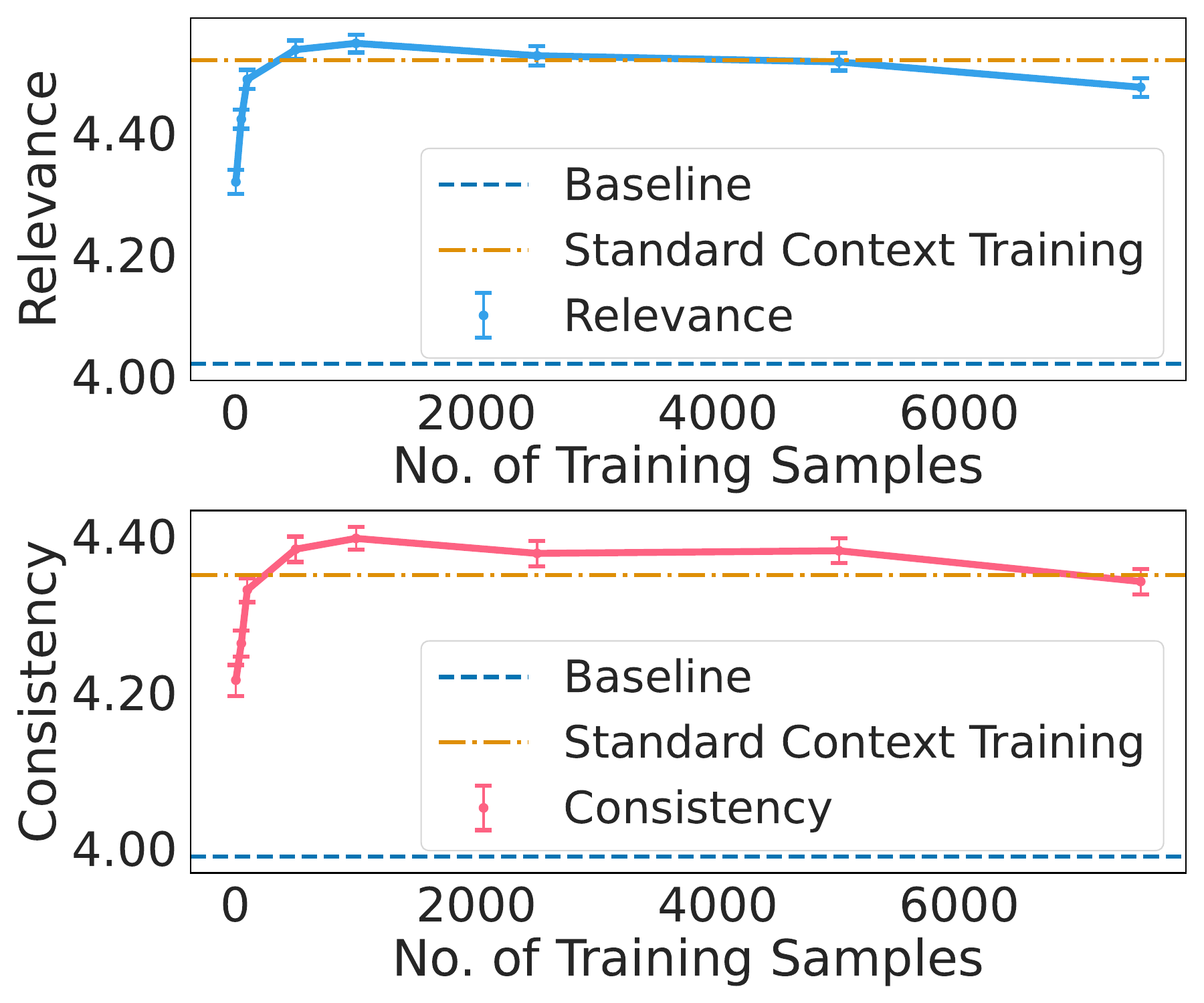}
    \vspace{1pt}
        
    \caption{Mixtral 8x7B}
    \end{subfigure}
    ~
        
        
\caption{SQuALITY: Relevance and consistency performance vs. number of synthetic training samples.}
        \label{fig:variable_data_squality}
\end{figure*}

\begin{figure*}
    \centering
    \begin{subfigure}[t]{0.32\textwidth}
    \centering
    \includegraphics[width=1\linewidth]{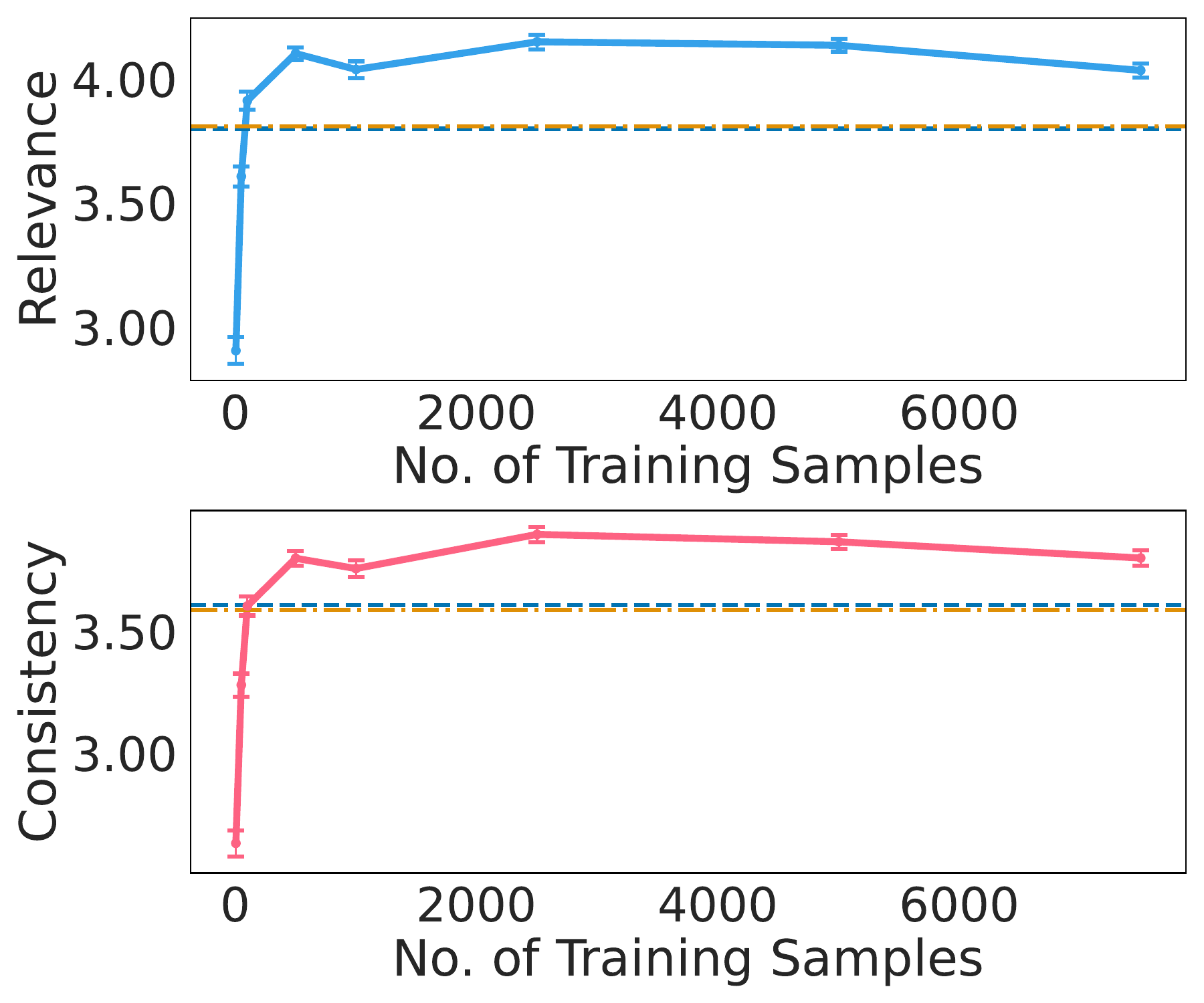}
    \vspace{1pt}
    
    \caption{Llama 3.2 1B}
    \end{subfigure}
    ~
    \begin{subfigure}[t]{0.32\textwidth}
        \centering\includegraphics[width=1\linewidth]{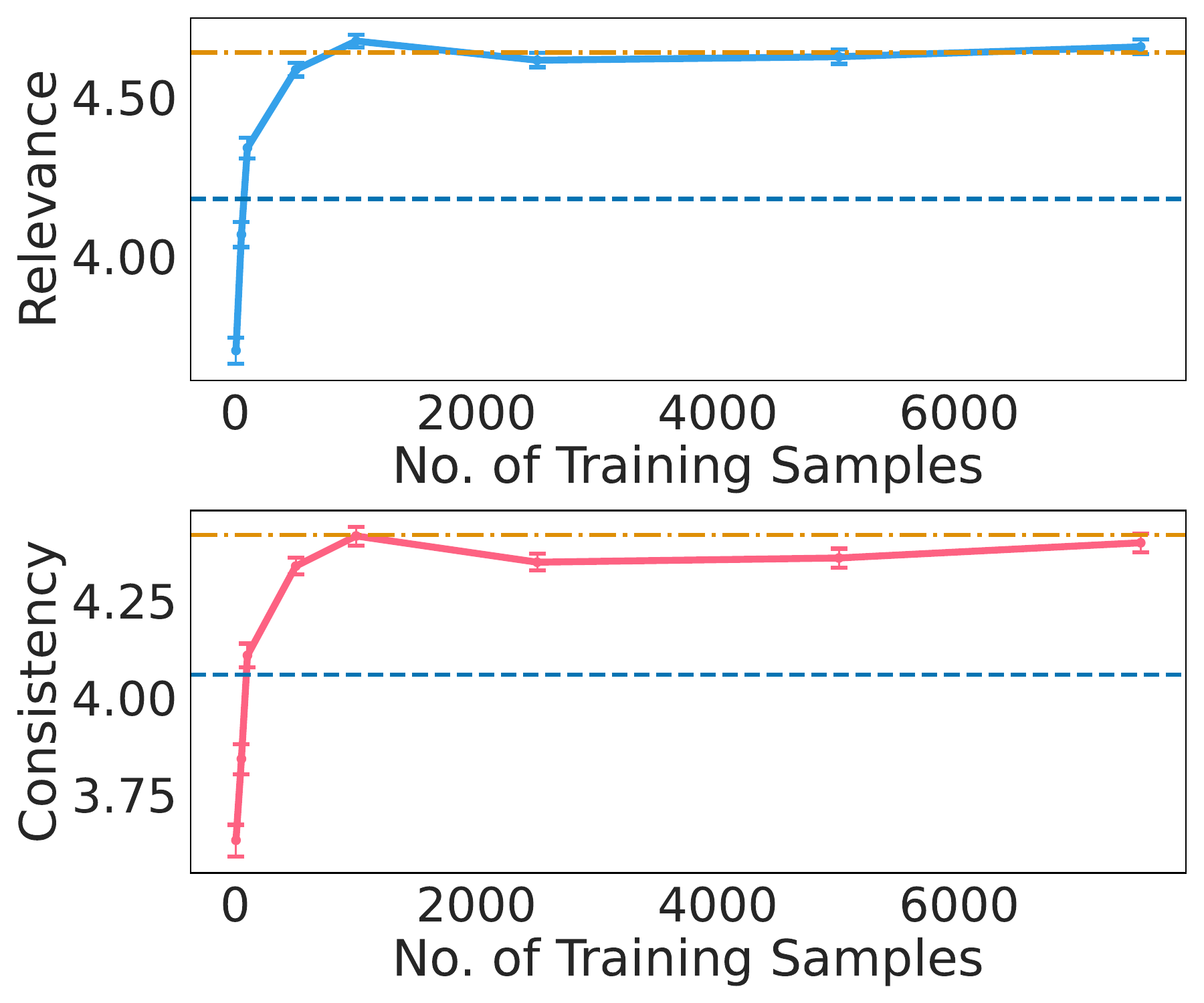}
    \vspace{1pt}
    
    \caption{Llama 3.1 8B}
    \end{subfigure}
    \begin{subfigure}[t]{0.32\textwidth}
        \centering\includegraphics[width=1\linewidth]{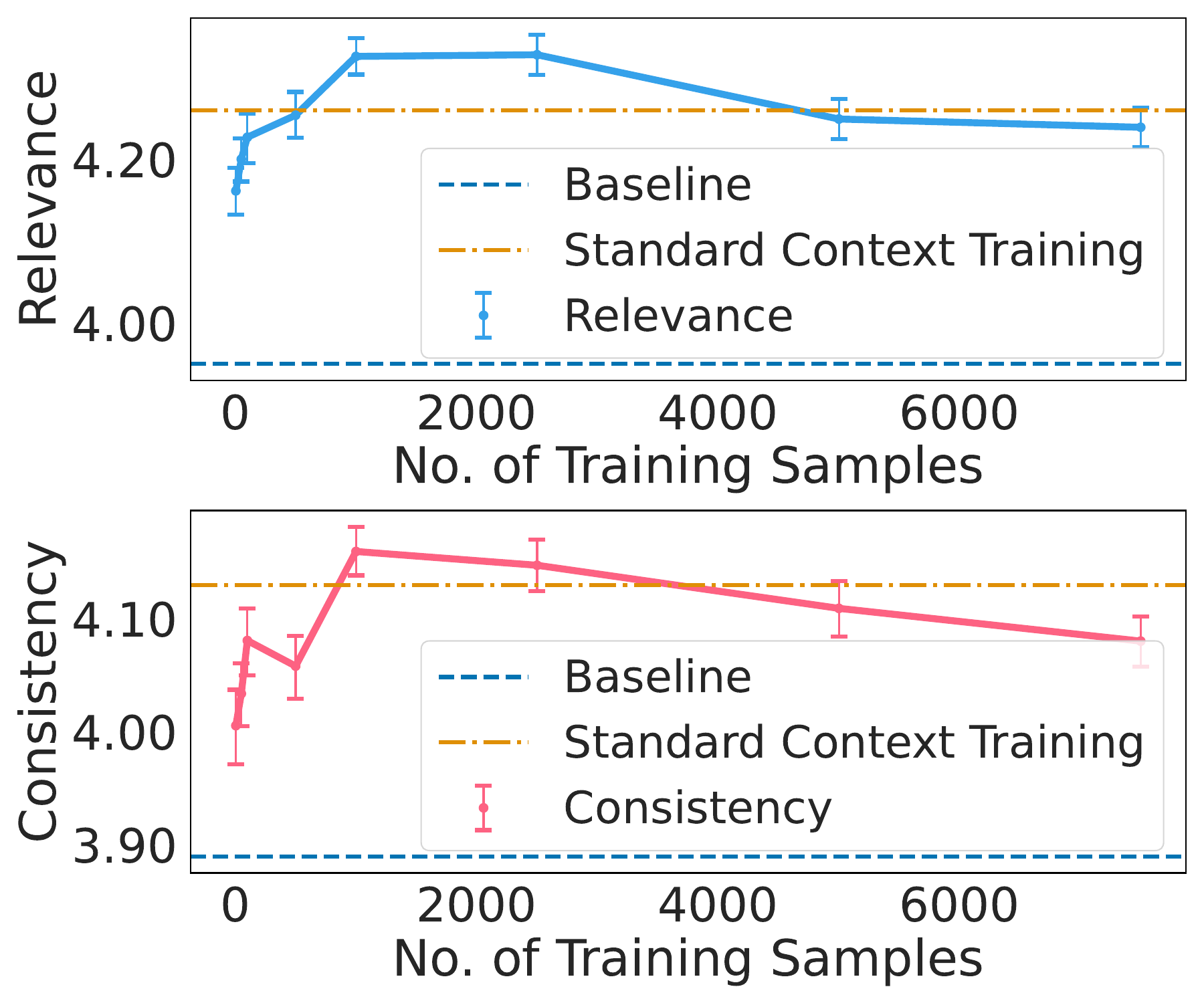}
    \vspace{1pt}
    
        \caption{Mixtral 8x7B}
    \end{subfigure}
    
    
\caption{ScholarQABench: Relevance and consistency performance vs. number of synthetic training samples.}
        \label{fig:variable_data_ScholarQABench}
\end{figure*}

To observe the impact of number of \dataset{} training samples on summary quality, we plot relevance and consistency vs. number of training samples for \sqa{} and \ops{} in \autoref{fig:variable_data_squality} and \autoref{fig:variable_data_ScholarQABench}. Interestingly, we find that performance generally peaks with only a modest amount of data (around 1k-3k samples depending on the model) at which point performance plateaus or slightly drops. It is likely that performance peaks when there is enough data to largely cover the distribution of data which is relevant for learning the task. Thus, more data does not result in more gains in performance, leading to the plateaus we see. We could potentially see additional performance gains by controlling the style of document generated, for example generating data which matches the target domain.

\section{Data Availability Statement}
We create \dataset{} in this work, as well as the code to generate \dataset{}, which we release freely to the public under the MIT license.\footnote{\url{https://github.com/dwright37/unstructured-evidence-sunset}}
The data are generated as sets of fiction and non-fiction books in English.

 \section{Model Descriptions}
 \label{sec:models_appendix}
Table \autoref{tab:models_description} presents the full set of Huggingface model identifiers for the LLMs used in our experiments. The model cards containing relevant information on number of parameters, context length, vocabulary size, etc. are available on their model page on the Huggingface website. All training and inference are performed using 1-2 Nvidia A100 GPUs with 48GB of memory. Prior to training we ran a brief hyperparameter search to find the parameters used in this study, sweeping over the following values (selected values in \textbf{bold}):
\begin{itemize}[noitemsep]
    \item Learning rate: $[$1e-6, 5e-4$]$ (\textbf{5e-5})
    \item Batch size: \{\textbf{2}, 4, 8, 16, 32\}
    \item Warmup steps: \{0, \textbf{10}, 50, 100, 150, 200, 300\}
    \item Train epochs: \{1, 2, 3, 4, 5, 8, \textbf{10}, 12, 20\}
    \item Lora rank: \{2, 4, 8, 12, \textbf{16}, 32\}
\end{itemize}

\begin{table}[t]
    \def\arraystretch{1.1}
    \centering
    \scriptsize
    \begin{tabular}{l|c}
    \toprule
    Model & Huggingface Identifier \\
    \midrule
       Llama 3.2 1B  & \texttt{meta-llama/Llama-3.2-1B-Instruct} \\
        Llama 3.2 3B & \texttt{meta-llama/Llama-3.2-3B-Instruct} \\
         Llama 3.1 8B & \texttt{meta-llama/Meta-Llama-3.1-8B-Instruct} \\ 
          Mistral Nemo 2407 & \texttt{mistralai/Mistral-Nemo-Instruct-2407} \\
          Mixtral 8x7B & \texttt{mistralai/Mixtral-8x7B-Instruct-v0.1} \\
          \bottomrule
    \end{tabular}
    \caption{Huggingface identifiers for models used in our experiments.}
    \label{tab:models_description}
\end{table}

\section{Software Package Parameters}
\begin{itemize}[noitemsep]
    \item NLTK~\cite{DBLP:conf/acl/Bird06}: We use the punkt sentence tokenizer for sentence tokenization
    \item VLLM: We use top $p$ sampling at 90\% with a temperature of 1. for inference. We set maximum new generated tokens to 2,000
    \item OpenAI GPT 4o Mini: We use top $p$ sampling at 90\% with a temperature of 1 for all prompts except title generation (temperature set to 1.2) and filtering (deterministic highest probability token output).
    \item DeepSeek-V3: We use top $p$ sampling at 90\% with a temperature of 1 for all prompts.
\end{itemize}

\section{Evaluation Robustness}
\label{sec:evaluation_robustness}
We use autoraters (i.e. LLM as a judge) for much of our evaluation. While we use a previously validated prompting and modeling setup~\cite{DBLP:journals/corr/abs-2410-23463}, we use DeepSeek-V3 as our autorater due to its high performance and low cost. We validated the robustness of DeepSeek-V3 as an autorater by taking a sample of 710 outputs summaries from our evaluation and re-evaluating them with GPT 4o Mini~\cite{DBLP:conf/emnlp/LiuIXWXZ23}. We measure the Pearson's R correlation between the ratings (2 ratings per summary) given by GPT 4o mini and DeepSeek-V3, finding a strong correlation of 73.29. This indicates the robustness of our evaluation which relies on DeepSeek-V3.

\end{document}